\documentclass[11pt]{article}

\usepackage[preprint]{acl}

\usepackage{times}
\usepackage{latexsym}

\usepackage[T1]{fontenc}

\usepackage[utf8]{inputenc}

\usepackage{microtype}

\usepackage{inconsolata}

\usepackage{graphicx}
\usepackage{amsmath,amssymb}
\usepackage{booktabs,array}
\usepackage{algorithm}
\usepackage{algpseudocode}
\usepackage{tikz}
\usetikzlibrary{arrows.meta,positioning,calc,fit,backgrounds}
\newcommand{\method}{RUSD}
\DeclareRobustCommand{\stepnum}[1]{\smash{\tikz[baseline=(n.base)]{\node[circle,fill=black!75,text=white,inner sep=0pt,minimum size=1.1em,font=\scriptsize\bfseries](n){#1};}}}

\title{Learning from Self-Generated Responses:\\Reliability--Uncertainty Self-Distillation for Reasoning}

\author{
  Jinhao Pan,
  Bowen Wei,
  Yuqing Zhou,
  Ziwei Zhu \\
  George Mason University\\
  \texttt{\{jpan23, bwei2, yzhou31, zzhu20\}@gmu.edu}
  }

\begin{document}
\maketitle
\begin{abstract}
Reference solutions and question-specific rubrics are not always available to guide the teacher in self-distillation. A large language model's own solution attempts offer an alternative, but their quality is difficult to assess in open-ended tasks without automatic correctness checks, and even correct attempts can involve uncertain token predictions. We propose Reliability--Uncertainty Self-Distillation (\method{}) to learn from these attempts. The same model serves as teacher and student: the teacher receives the question and completed attempt, and the student matches the teacher's token-level predictions using a question-only prompt. \method{} weights each attempt's distillation loss by additively combining judged reliability, estimated from a self-judge's score and confidence, with student predictive uncertainty. Across eleven reasoning benchmarks, \method{} outperforms all five baselines in average accuracy, reaching 70.47\% on Qwen3-4B and 76.69\% on Qwen3-8B (8.90 and 2.10 percentage points above the base models). Data and code are available at \url{https://github.com/JP-25/RUSD}.
\end{abstract}

\section{Introduction}
\label{sec:introduction}

Self-distillation trains large language models (LLMs) by using the model itself as a teacher \citep{zhao2026selfdistilledreasoneronpolicyselfdistillation,gu2026rethinkingrewardsupervisionrubricconditioned}. In recent approaches, the teacher receives the question together with additional \emph{teacher context}, such as a reference solution; the student uses a question-only prompt and learns by matching the teacher's next-token predictions along student-generated responses.

\paragraph{Challenge 1: Reference solutions and rubrics are not always available to guide the teacher.}
SDFT and OPSD supply demonstration responses (example answers) or reference solutions as teacher context \citep{shenfeld2026selfdistillationenablescontinuallearning,zhao2026selfdistilledreasoneronpolicyselfdistillation}. RCSD and RGSD use question-specific rubrics, which specify what a good response should contain \citep{gu2026rethinkingrewardsupervisionrubricconditioned,rezaei2026rubricguidedselfdistillationposttrainingrubric}. When these solutions or rubrics are unavailable, preparing them can require expensive human annotation or additional model generation. One alternative is to learn from the model's own generated responses without supplied reference solutions \citep{pei2026negativeselfdistillationlearningreason,li2026onpolicyselfdistillationanysupervision}. These self-generated \emph{solution attempts}, however, may contain errors.

\paragraph{Challenge 2: Assessing attempt quality is difficult in open-ended reasoning.}
On open-ended reasoning tasks, reliable automatic verifiers for these self-generated attempts are often unavailable \citep{ye2026llmasacoachexperientiallearningnonverifiable}. In reinforcement learning with verifiable rewards (RLVR), mathematical answers can be checked against known results, and code can be run against tests \citep{deepseekai2025deepseekr1}. Without comparable checks or reference answers, it is difficult to determine which attempts are correct and how much to learn from them. An LLM \emph{judge}, a model prompted to assess response quality, can score these attempts, but its ratings can be wrong \citep{zheng2023judgingllmasajudgemtbench}. And using a larger external judge can increase the computation or API fees needed to score training attempts \citep{verga2024replacingjudgesjuriesevaluating,xie2026smalljudgesrubricrl}.

\begin{figure*}[t]
    \centering
    \includegraphics[width=\textwidth]{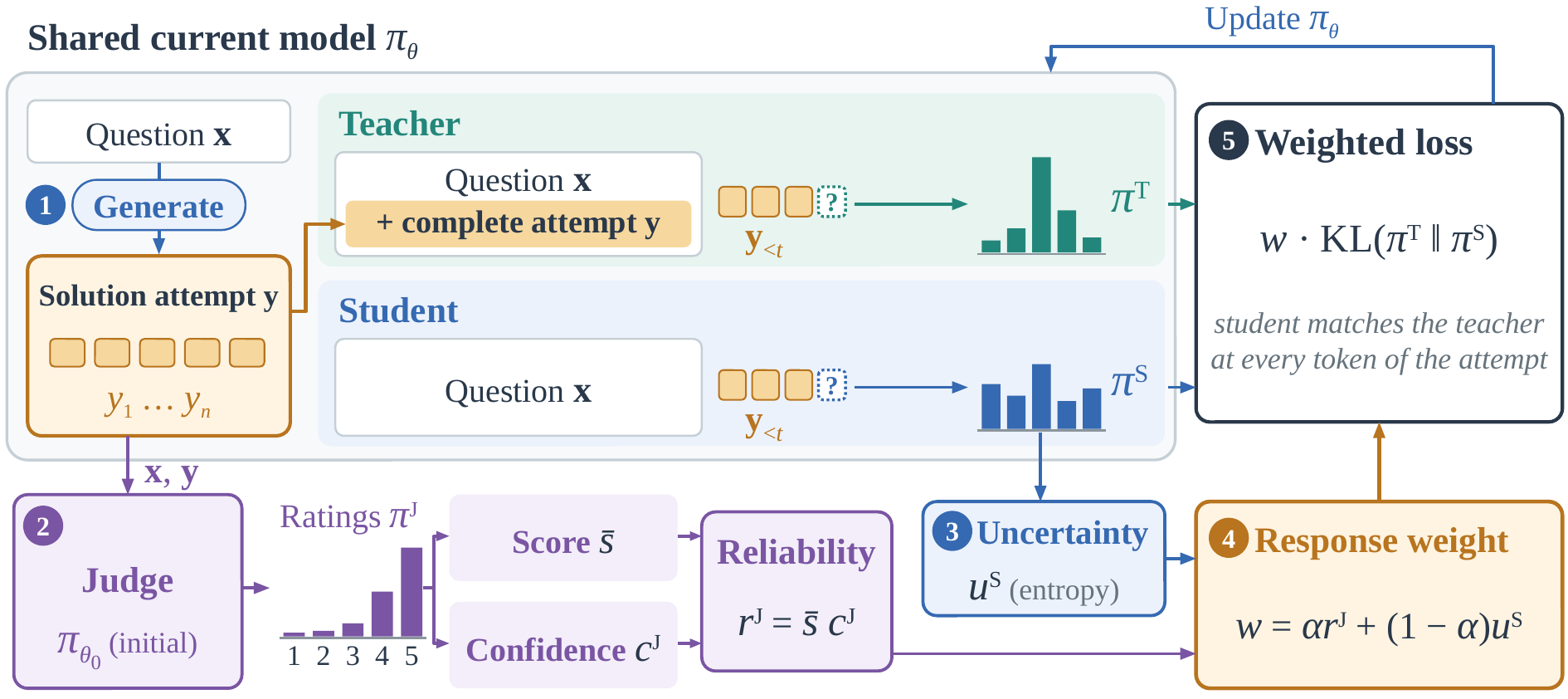}
    \caption{Overview of \method{}. \stepnum{1}~$\pi_\theta$ generates attempt $\mathbf y$ for question $\mathbf x$. Teacher and student predict next-token distributions from the same response prefix $\mathbf y_{<t}$ under different prompts; only the teacher prompt includes the complete attempt. The teacher generates no additional response. \stepnum{2}~Judge score $s$, normalized to $\bar s$, and confidence $c^{\mathrm J}$ give judged reliability $r^{\mathrm J}$. \stepnum{3}~Normalized student entropy averaged over the response gives uncertainty $u^{\mathrm S}$. \stepnum{4}~A weighted sum of reliability and uncertainty gives response weight $w$. \stepnum{5}~The weighted distillation loss updates the shared model. Numbers match Sections~\ref{sec:context}--\ref{sec:loss}; bars are schematic.}
    \label{fig:overview}
\end{figure*}

Our intuition is to learn from correct attempts generated with uncertain token predictions, so the model can produce the reasoning more confidently. 
We propose \emph{Reli\-ability--\allowbreak Un\-certainty Self-Distillation} (\method{}, Figure~\ref{fig:overview}) based on this intuition. For each question, the current model generates one solution attempt. As teacher, the same model receives the question and completed attempt in its prompt to provide token-level guidance. A judge assesses the attempt's quality: high scores supported by confident judgments indicate greater \emph{judged reliability}. As student, the model learns to match the teacher's guidance with only the question as its prompt. We weight each attempt's distillation loss using judged reliability and the student's uncertainty in predicting its tokens, giving priority to attempts judged reliable and to attempts with uncertain token predictions.

Our contributions are threefold: \textbf{1.} We introduce \method{}, which uses one model-generated attempt as teacher context without supplied solutions or question-specific rubrics. \textbf{2.} We design response weights that combine judged reliability and student uncertainty, giving greater learning priority to attempts judged reliable and to attempts with uncertain token predictions. \textbf{3.} Across eleven reasoning benchmarks, \method{} outperforms all five baselines in macroaverage accuracy at both scales, improving Qwen3-4B and Qwen3-8B by 8.90 and 2.10 over the base models. External ratings, training analyses, and ablations examine its weighting signals and learning behavior.

\section{Related Work}
\label{sec:related}

\paragraph{Self-distillation with and without supplied references.}
Self-distillation trains a model to match its own predictions. In recent on-policy LLM methods, the teacher receives additional context and the student matches the teacher's token distributions along student-generated responses \citep{zhao2026selfdistilledreasoneronpolicyselfdistillation}. SDFT and OPSD supply demonstrations or reference solutions; RCSD and RGSD supply question-specific rubrics \citep{shenfeld2026selfdistillationenablescontinuallearning,zhao2026selfdistilledreasoneronpolicyselfdistillation,gu2026rethinkingrewardsupervisionrubricconditioned,rezaei2026rubricguidedselfdistillationposttrainingrubric}. Without supplied solutions, NSD constructs negative teacher conditions from model responses, whereas Canon and u-OPSD select a self-generated solution by majority-answer consensus across multiple responses \citep{pei2026negativeselfdistillationlearningreason,gkountouras2026consensusprivilegedcontextlabelfree,li2026onpolicyselfdistillationanysupervision}. \method{} uses one completed attempt as teacher context, without negative conditions or answer consensus.

\paragraph{Assessing self-generated attempts.}
SDPO supplies environment feedback or successful responses to the teacher, while SD-Zero trains a reviser with binary correctness feedback \citep{hubotter2026reinforcementlearningselfdistillation,he2026selfdistillationzeroselfrevisionturns}. Such feedback requires response assessment; consensus requires rules for matching answers. \method{} estimates attempt reliability from judge score and confidence without correctness labels. Reliability controls the response weight; the teacher supplies next-token predictions.

\paragraph{Weighting by confidence and uncertainty.}
LLKD's weighted variant combines teacher confidence and student uncertainty for classification \citep{li-etal-2025-learning}. EGRSD weights token updates using teacher entropy, PW-OPSD uses token position, and COSE uses evaluator confidence for policy updates and replay \citep{ke2026respectingselfuncertaintyonpolicyselfdistillation,liu2026whenteachertokensreliable,wei2026confidenceorchestratedselfevolutionuncertainllm}. \method{} combines judged reliability with student predictive uncertainty to weight each attempt's distillation loss. Appendix~\ref{sec:related-extended} gives fuller comparisons and the scope of our baselines.

\section{Method}
\label{sec:method}

\paragraph{Overview.}
Figure~\ref{fig:overview} illustrates the \method{} workflow, and Algorithm~\ref{alg:acsd} in Appendix~\ref{sec:implementation} shows the training procedure. \method{} trains on questions $\mathcal D=\{\mathbf{x}_i\}$ without supplied reference solutions or question-specific rubrics. The current model $\pi_\theta$ serves as teacher and student under different prompts. The initial model $\pi_{\theta_0}$ is kept frozen throughout training and serves as the judge. The current model generates one attempt, which the teacher uses as additional context to provide token-level guidance (\stepnum{1}). The judge's score and confidence estimate the attempt's reliability (\stepnum{2}). We also measure the student's uncertainty in predicting the attempt's tokens (\stepnum{3}). These signals determine the attempt's training weight (\stepnum{4}). The student learns to match the teacher's token distributions, with each attempt's contribution scaled by its weight (\stepnum{5}). Table~\ref{tab:prompts} in Appendix~\ref{sec:implementation} gives the three prompts.

\subsection{Using the attempt as teacher context}
\label{sec:context}

For each sampled question $\mathbf{x}_i$, the current model $\pi_\theta$ generates one solution attempt $\mathbf y_i$:
\begin{equation}
    \mathbf{y}_i\sim\pi_\theta(\cdot\mid\mathcal P_{\mathrm S}(\mathbf{x}_i)),
    \label{eq:rollout}
\end{equation}
where $\mathcal P_{\mathrm S}$ is the question-only prompt. The teacher prompt $\mathcal P_{\mathrm T}(\mathbf{x}_i,\mathbf{y}_i)$ contains the question and completed attempt. The prompt describes the attempt as potentially useful or misleading and asks the model to solve the problem independently using any useful information.

Following on-policy self-distillation \citep{zhao2026selfdistilledreasoneronpolicyselfdistillation}, we evaluate teacher and student next-token distributions at every position within this same sampled response. At position $t$, both receive the preceding response tokens $\mathbf{y}_{i,<t}=(y_{i,1},\ldots,y_{i,t-1})$. Only the teacher prompt also contains a complete copy of $\mathbf y_i$, giving it access to the response's later reasoning when predicting earlier tokens:
\begin{align}
    \pi^{\mathrm S}_{i,t}(v)&=\pi_\theta(v\mid\mathcal P_{\mathrm S}(\mathbf{x}_i),\mathbf{y}_{i,<t}),\nonumber\\
    \pi^{\mathrm T}_{i,t}(v)&=\pi_\theta(v\mid\mathcal P_{\mathrm T}(\mathbf{x}_i,\mathbf{y}_i),\mathbf{y}_{i,<t}),
    \label{eq:aligned}
\end{align}
where $v\in\mathcal V$ is a vocabulary token, and $\pi^{\mathrm S}_{i,t}$ and $\pi^{\mathrm T}_{i,t}$ are the student and teacher next-token distributions, respectively.

\subsection{Estimating reliability from self-judgment}
\label{sec:judge}

The judge $\pi_{\theta_0}$ assesses each completed attempt to provide a quality signal for response weighting; the teacher provides token distributions for the student to match. For efficiency, we reuse the initial model as judge rather than rely on a larger external model. Given the question and attempt, the judge evaluates logical validity, internal consistency, factual plausibility, relevance, and completeness using one five-level rubric shared across questions (Table~\ref{tab:prompts}, Appendix~\ref{sec:implementation}).

For question--attempt pair $i$, let $\ell^{\mathrm J}_{i,k}$ denote the judge's logit for rating $k\in\{1,\ldots,5\}$; the superscript $\mathrm J$ labels judge quantities. The judge is prompted to return one rating digit. We read its five rating-token logits and normalize them into a distribution, with $j$ indexing the possible ratings in the denominator:
\begin{equation}
    \pi^{\mathrm J}_i(k)=\frac{\exp(\ell^{\mathrm J}_{i,k})}
    {\sum_{j=1}^{5}\exp(\ell^{\mathrm J}_{i,j})}.
    \label{eq:judge-probs}
\end{equation}
\paragraph{Judge score.}
We take the judge's most probable rating as the quality score: 1 means fundamentally incorrect, and 5 means fully correct and well justified. We rescale the score to $[0,1]$ for use in the response weight:
\begin{equation}
    s_i=\arg\max_{k\in\{1,\ldots,5\}}\pi^{\mathrm J}_i(k),
    \qquad \bar s_i=\frac{s_i-1}{4}.
    \label{eq:score}
\end{equation}

\paragraph{Judge confidence.}
Judge confidence distinguishes attempts with the same score: the judge may strongly favor that rating for one attempt but be split between ratings for another. We quantify this distinction using one minus normalized rating entropy:
\begin{equation}
    c_i^{\mathrm J}=1-\frac{H(\pi^{\mathrm J}_i)}{\log5},
    \label{eq:judge-confidence}
\end{equation}
where $H(p)=-\sum_jp(j)\log p(j)$ is Shannon entropy and $\log5$ is the maximum entropy over five ratings. Confidence is zero for equally likely ratings and approaches one as probability concentrates on one rating. It measures decisiveness, not a calibrated probability of correctness.

\paragraph{Judged reliability.}
We combine the normalized judge score and confidence into judged reliability:
\begin{equation}
    r_i^{\mathrm J}=\bar s_i c_i^{\mathrm J}.
    \label{eq:quality}
\end{equation}
The product is high only when the judge assigns a high score with high confidence; uncertain high scores and confident low scores both receive lower reliability. Appendix~\ref{sec:implementation} gives an example, and Section~\ref{sec:judge-audit} relates judge score and confidence to externally assessed response quality.

\subsection{Measuring student uncertainty}
\label{sec:uncertainty}

We use student uncertainty to prioritize attempts whose tokens the student predicts less confidently, including attempts judged reliable. We measure this uncertainty by averaging normalized next-token entropy over the generated response $\mathbf y_i$ of length $n_i=|\mathbf{y}_i|$:
\begin{equation}
    u_i^{\mathrm S}=\frac{1}{n_i}\sum_{t=1}^{n_i}
        \frac{H(\pi^{\mathrm S}_{i,t})}{\log|\mathcal V|}.
    \label{eq:uncertainty}
\end{equation}
We use the full-vocabulary student distribution at the distillation temperature (Appendix~\ref{sec:implementation}); prompt and padding positions are excluded. Normalization bounds $u_i^{\mathrm S}$ in $[0,1]$, and averaging avoids an automatic increase with response length. A larger value indicates more diffuse token predictions along that response, rather than a direct estimate of whether its reasoning is correct.

\subsection{Combining reliability and uncertainty}
\label{sec:weight}

We combine the two signals to weight each attempt's distillation loss:
\begin{equation}
    w_i=\alpha r_i^{\mathrm J}+(1-\alpha)u_i^{\mathrm S},
    \qquad \alpha\in[0,1].
    \label{eq:weight}
\end{equation}
The reliability term prioritizes attempts judged to be high quality with high confidence, while the uncertainty term prioritizes attempts for which the student's token predictions are uncertain.

The additive rule lets either priority contribute independently: uncertainty can contribute when judged reliability is low, and reliability can contribute when the student is already certain. The endpoints $\alpha=1$ and $\alpha=0$ use reliability and uncertainty alone, respectively. We use $\alpha=0.5$, which gives both signals equal coefficients; their realized contributions depend on their magnitudes. Section~\ref{sec:ablations} evaluates both components and their balance.

\subsection{Training objective}
\label{sec:loss}

We train the student to match the teacher's token distributions, scaling each response's average forward-KL loss by its weight $w_i$:
\begin{equation}
    \mathcal L(\theta)=\mathbb E_{\mathbf{x}_i,\mathbf{y}_i}\!\left[
       \frac{w_i}{n_i}\sum_{t=1}^{n_i}
       D_{\mathrm{KL}}(\pi^{\mathrm T}_{i,t}\,\|\,\pi^{\mathrm S}_{i,t})
    \right],
    \label{eq:loss}
\end{equation}
where the expectation is over questions from $\mathcal D$ and attempts drawn from Equation~\ref{eq:rollout}. Following OPSD \citep{zhao2026selfdistilledreasoneronpolicyselfdistillation}, we optimize a coordinate-clipped surrogate of this loss, with gradients passing only through student predictions. Appendix~\ref{sec:implementation} gives the clipping and model-update details.

\section{Experiments}
\label{sec:experiments}

\begin{table*}[t]
\centering
\fontsize{9}{10.8}\selectfont
\setlength{\tabcolsep}{2pt}
\renewcommand{\arraystretch}{1.0}

\begin{tabular*}{\textwidth}{@{\extracolsep{\fill}}l*{12}{r}@{}}
\toprule
& \multicolumn{5}{c}{Science and general reasoning}
& \multicolumn{5}{c}{Mathematical reasoning}
& \multicolumn{1}{c}{Logic}
& \multicolumn{1}{c}{Overall} \\
\cmidrule(lr){2-6}\cmidrule(lr){7-11}\cmidrule(lr){12-12}\cmidrule(l){13-13}
Method & GPQA & SciB & MMLU & LiveB & OBQA & GSM8K & MATH & AMC & Minerva & Olymp & Zebra & Average \\
\midrule
\multicolumn{13}{@{}l}{\textbf{Qwen3-4B}} \\
\addlinespace[2pt]
Base & 41.16 & 35.77 & 78.12 & 46.50 & 91.40 & 93.03 & 81.68 & 62.35 & 29.04 & 46.03 & 72.13 & 61.56 \\
SFT\textsuperscript{\textdagger} & 42.30 & 35.91 & 78.19 & 42.00 & 91.00 & 93.18 & 80.50 & 54.22 & 29.41 & 44.51 & 72.10 & 60.30 \\
GRPO\textsuperscript{\textdagger} & 43.81 & 43.89 & 78.85 & 47.00 & \textbf{93.00} & 93.78 & 81.86 & 63.86 & 31.25 & 47.92 & 72.58 & 63.44 \\
RCSD\textsuperscript{\textdagger} & 41.41 & 42.56 & 78.66 & 49.50 & 92.80 & 93.25 & 81.64 & 61.45 & 30.15 & 45.99 & 71.53 & 62.63 \\
OPSD\textsuperscript{\textdagger} & 51.14 & 46.32 & 77.49 & 47.50 & 91.60 & 94.24 & 84.48 & 71.08 & 29.04 & 52.67 & 69.15 & 64.97 \\
NSD & 51.14 & 61.81 & 79.22 & 47.62 & \textbf{93.00} & 94.22 & 81.83 & 62.65 & 29.78 & 45.96 & 72.50 & 65.43 \\
\method{} & \textbf{52.78} & \textbf{70.66} & \textbf{79.38} & \textbf{61.25} & 92.60 & \textbf{94.81} & \textbf{85.87} & \textbf{74.10} & \textbf{35.02} & \textbf{54.27} & \textbf{74.38} & \textbf{70.47} \\
\midrule
\multicolumn{13}{@{}l}{\textbf{Qwen3-8B}} \\
\addlinespace[2pt]
Base & 62.12 & 75.72 & 82.04 & 70.12 & 93.70 & 96.17 & 87.20 & 80.12 & 34.10 & 58.05 & 81.13 & 74.59 \\
SFT\textsuperscript{\textdagger} & 61.24 & 74.49 & 82.95 & 67.62 & 93.85 & 96.17 & 87.13 & 80.42 & 34.65 & 58.42 & 82.70 & 74.51 \\
GRPO\textsuperscript{\textdagger} & 62.63 & 75.54 & 82.95 & 72.50 & 94.40 & 96.10 & 88.20 & 81.33 & 34.28 & 59.27 & 82.93 & 75.47 \\
RCSD\textsuperscript{\textdagger} & 60.61 & 76.02 & 83.00 & 70.62 & 95.60 & 95.98 & 87.95 & 81.33 & 34.19 & 59.79 & 77.25 & 74.76 \\
OPSD\textsuperscript{\textdagger} & 59.72 & 76.59 & 80.18 & 70.63 & 94.10 & 95.81 & 88.31 & 80.12 & 33.92 & 60.42 & 81.25 & 74.64 \\
NSD & 64.65 & 74.96 & 83.06 & 72.50 & 95.25 & 96.23 & 88.06 & 82.53 & 34.01 & 59.53 & 82.93 & 75.79 \\
\method{} & \textbf{64.90} & \textbf{77.13} & \textbf{83.58} & \textbf{73.13} & \textbf{95.70} & \textbf{97.52} & \textbf{88.48} & \textbf{83.13} & \textbf{36.03} & \textbf{60.87} & \textbf{83.12} & \textbf{76.69} \\
\bottomrule
\end{tabular*}
\caption{Accuracy (Avg@4, \%) on eleven reasoning benchmarks. \method{} achieves the highest macroaverage at both model sizes. Average is the unweighted mean over benchmarks; bold marks the best score in each column within a model size, including ties. \textsuperscript{\textdagger}Uses supervision beyond the training questions and the model's own outputs: dataset responses (SFT, OPSD), generated rubrics (RCSD), or an external LLM judge (GRPO). GPQA: GPQA Diamond; SciB: SciBench; LiveB: LiveBench Reasoning; OBQA: OpenBookQA; Minerva: Minerva Math; Olymp: OlympiadBench; Zebra: ZebraLogic.}
\label{tab:main}
\end{table*}

Our experiments examine reasoning performance, the judge's quality signals, training dynamics, and design choices: \textbf{RQ1:} Can \method{} improve reasoning accuracy across tasks and model scales? \textbf{RQ2:} Do judge scores and confidence reflect response quality? \textbf{RQ3:} During training, does the number of responses with high judge scores and confidence increase, and does the percentage with low student uncertainty within this group also increase? \textbf{RQ4:} How do the weighting signals, their balance, and the loss type affect performance?

\subsection{Experimental Setup}
\label{sec:setup}

\paragraph{Models and Training Data.}
We evaluate Qwen3-4B and Qwen3-8B \citep{qwen2025qwen3}, with component ablations and hyperparameter studies on 4B. All methods share 30,000 NaturalReasoning questions \citep{yuan2025naturalreasoningreasoningwild28m}, sampled with seed 42 after filtering for non-empty reference answers so that reference-based baselines can use the same questions. \method{} learns from questions and its own online attempts; the dataset's reference answers and generated responses are excluded from its teacher and judge prompts.

\paragraph{Baselines.}
We compare five representative supervision strategies on the same training questions. \emph{Supervised fine-tuning (SFT)} \citep{ouyang2022traininglanguagemodelsfollow} learns from the dataset's Llama3.3-70B-Instruct responses. \emph{GRPO} \citep{shao2024deepseekmath} uses scalar rewards from an external LLM judge. \emph{RCSD} \citep{gu2026rethinkingrewardsupervisionrubricconditioned} supplies question-specific rubrics as teacher context. \emph{OPSD} \citep{zhao2026selfdistilledreasoneronpolicyselfdistillation} uses the dataset's Llama3.3-70B-Instruct responses as teacher context and matches teacher and student predictions along student-generated responses. \emph{NSD} \citep{pei2026negativeselfdistillationlearningreason} generates negative conditions from model responses and suppresses behavior encouraged by the negatively conditioned teacher. Thus SFT, OPSD, RCSD, and GRPO rely on supervision beyond the training questions (marked with \textsuperscript{\textdagger} in Table~\ref{tab:main}), whereas NSD and \method{} use only the questions and the model's own outputs.

\paragraph{Benchmarks.}
We evaluate eleven benchmarks: GPQA Diamond \citep{rein2024gpqa}, SciBench \citep{wang2024scibench}, MMLU \citep{hendrycks2021measuring}, LiveBench Reasoning \citep{white2025livebench}, and OpenBookQA \citep{mihaylov-etal-2018-suit} for scientific and general reasoning; GSM8K \citep{cobbe2021trainingverifiers}, MATH \citep{hendrycks2021measuringmathematical}, AMC \citep{aimo2024amc}, Minerva Math \citep{lewkowycz2022solving}, and OlympiadBench \citep{he-etal-2024-olympiadbench} for mathematics; and ZebraLogic \citep{lin2025zebralogic} for logic-grid reasoning.

\paragraph{Evaluation.}
We report Avg@4: correctness averaged over four generated responses per question and then across questions, expressed as a percentage. \emph{Average} in Table~\ref{tab:main} is the unweighted mean of the eleven benchmark scores. All accuracy gains are absolute differences on this 0--100 scale, calculated before rounding. Appendix~\ref{sec:implementation} gives the decoding and scoring settings.

\paragraph{Implementation.}
All methods use NVIDIA A100 80\,GB GPUs and shared training settings. At both model sizes, results are reported after 150 training steps, except GRPO, which uses 500 steps as in RCSD's GRPO setting. For \method{}, we train LoRA adapters \citep{hu2021loralowrankadaptationlarge} on four GPUs with $\alpha=0.5$ and student and teacher prompts in thinking mode. Appendix~\ref{sec:implementation} gives the hyperparameters (Table~\ref{tab:hyperparameters}), model updates, GRPO judge, and evaluation-step details.

\subsection{Reasoning Performance (RQ1)}
\label{sec:main-results}

\begin{figure*}[t]
\centering
\begin{minipage}{0.33\textwidth}
\centering
\includegraphics[width=\linewidth,trim=0bp 5bp 0bp 4bp,clip]{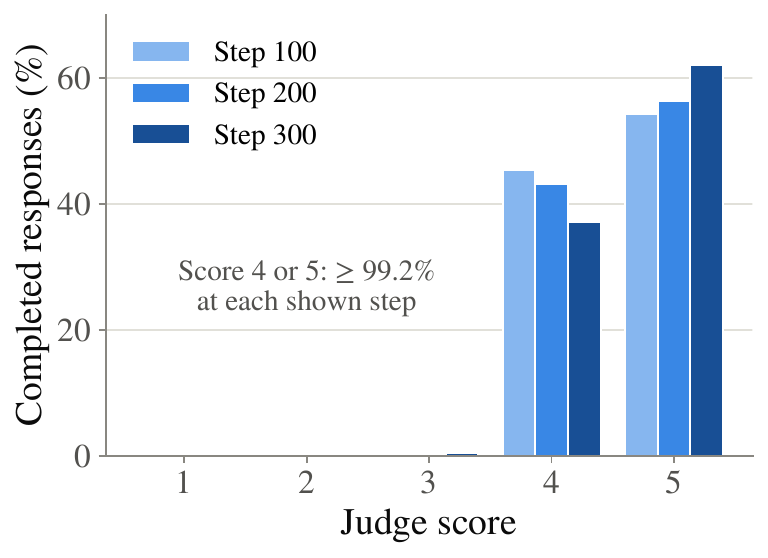}
\small (a) Judge scores
\end{minipage}\hfill
\begin{minipage}{0.33\textwidth}
\centering
\includegraphics[width=\linewidth,trim=0bp 5bp 0bp 4bp,clip]{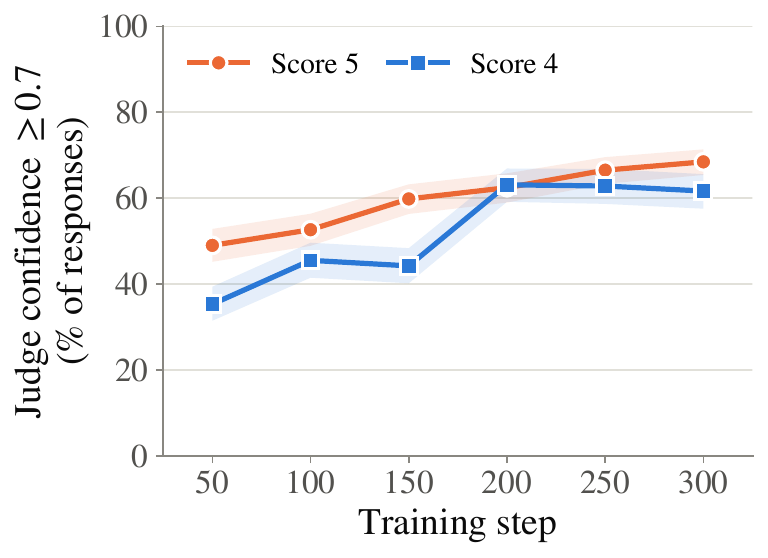}
\small (b) High-confidence judgments
\end{minipage}\hfill
\begin{minipage}{0.33\textwidth}
\centering
\includegraphics[width=\linewidth,trim=0bp 5bp 0bp 4bp,clip]{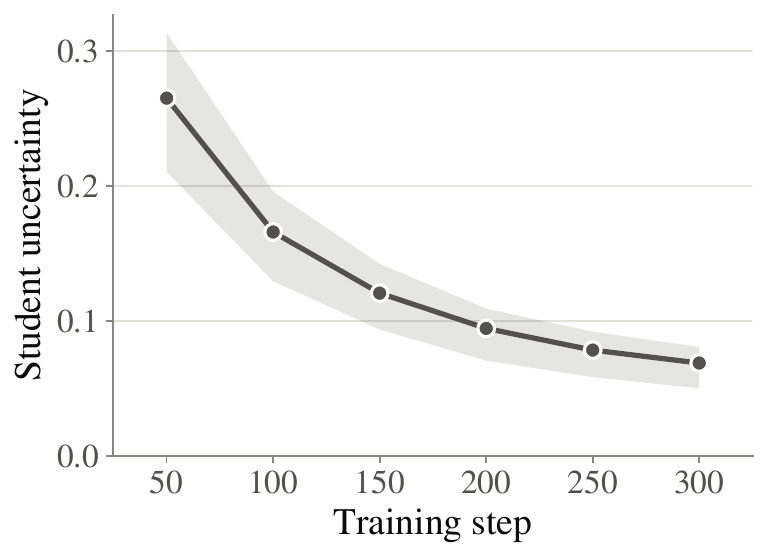}
\small (c) Student uncertainty
\end{minipage}
\caption{Judge assessments and student uncertainty during Qwen3-4B \method{} training with $\alpha=0.5$. \textbf{(a)} Distribution of judge scores over completed responses in summaries reported at training steps 100, 200, and 300. \textbf{(b)} Percentage of high-confidence judgments ($c_i^{\mathrm J}\geq0.7$) within the score-4 and score-5 categories; shading shows 95\% Wilson confidence intervals for these proportions. \textbf{(c)} Mean student uncertainty over completed score-4/5 responses, with shading spanning the 25th--75th percentiles.}
\label{fig:dynamics}
\end{figure*}

\paragraph{\method{} outperforms all five baselines in average accuracy at both model scales.}
In Table~\ref{tab:main}, \method{} reaches 70.47\% macroaverage accuracy at 4B and 76.69\% at 8B, exceeding the base models by 8.90 and 2.10 and NSD, the strongest evaluated baseline by average, by 5.04 and 0.90, respectively. It leads on ten of eleven benchmarks at 4B and all eleven at 8B, with gains spanning scientific and general reasoning, mathematics, and logic-grid reasoning. At 4B, notable results include 70.66\% on SciBench versus NSD's 61.81\%, and 61.25\% on LiveBench Reasoning, 11.75 above the best baseline (RCSD). OpenBookQA is the exception: 92.60\% versus 93.00\% for GRPO and NSD. At 8B, gains over the best baseline on each benchmark range from 0.10 to 1.38, despite the stronger starting model.

The gains also hold against external supervision: at 4B, \method{} exceeds OPSD by 5.49, while SFT on the same supplied solutions falls below the base model. At 8B, the four externally supervised baselines change the base average by $-$0.08 to +0.88, compared with \method{}'s +2.10. Thus, \method{} improves both tested models without supplied solutions or question-specific rubrics. Every evaluated 4B training step from 50 to 300 also exceeds the strongest baseline average (Appendix~\ref{sec:training-duration}).

\begin{figure*}[t]
\centering
\begin{minipage}{0.45\textwidth}
\centering
\includegraphics[width=\linewidth]{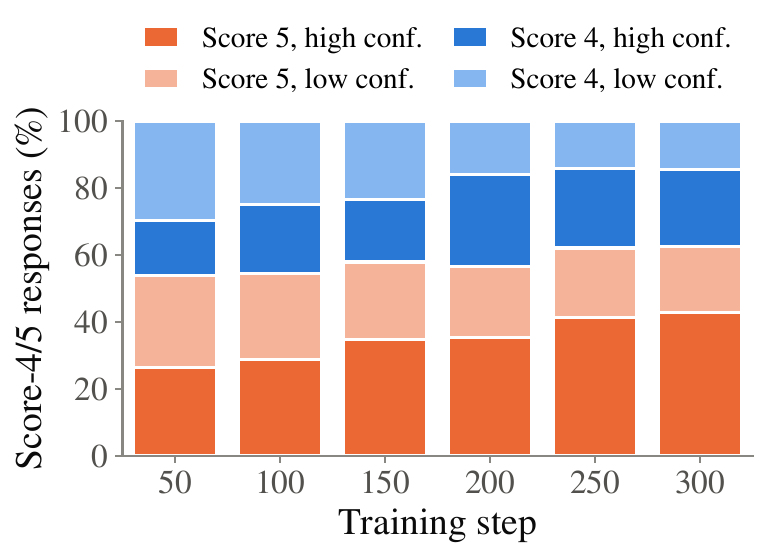}
\small (a) Judge-score/confidence groups
\end{minipage}\hfill
\begin{minipage}{0.45\textwidth}
\centering
\includegraphics[width=\linewidth]{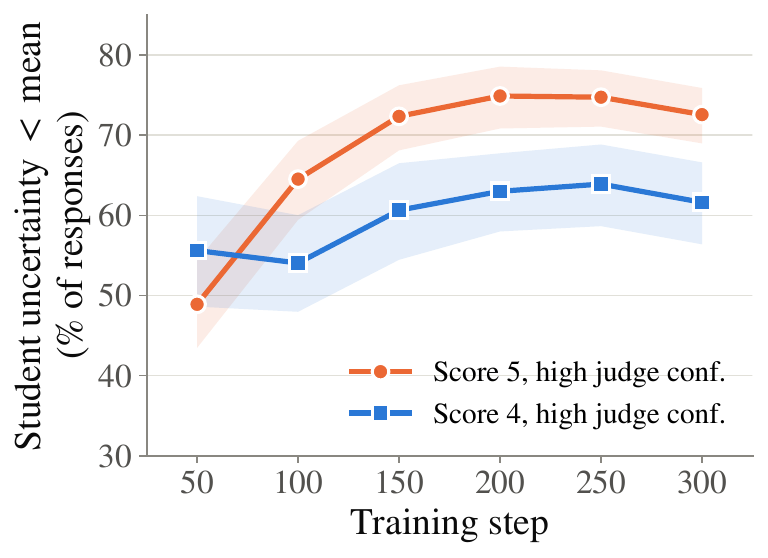}
\small (b) Low student uncertainty by group
\end{minipage}
\caption{Response-group composition and low student uncertainty during Qwen3-4B \method{} training with $\alpha=0.5$. High judge confidence means $c_i^{\mathrm J}\geq0.7$. \textbf{(a)} Shares of completed score-4/5 responses in four groups defined by judge score and confidence. \textbf{(b)} Percentage of responses within each high-confidence group whose student uncertainty is below the pooled mean of all completed score-4/5 responses at that reporting step. Shading shows 95\% Wilson confidence intervals from group response counts. Both panels use the same response population at each reporting step (Appendix~\ref{sec:implementation}).}
\label{fig:student-groups}
\end{figure*}

\subsection{Judge Scores, Confidence, and Response Quality (RQ2)}
\label{sec:judge-audit}

\begin{table}[t]
\centering
\small
\setlength{\tabcolsep}{3pt}
\begin{tabular}{@{}l r r r@{}}
\toprule
\textbf{Predictors} & Multiple $R$ & $R^2$ (\%) & $p$-value\\
\midrule
Judge score & 0.46 & 21 & $8.9\times10^{-12}$\\
Judge confidence & 0.62 & 39 & $5.0\times10^{-23}$\\
Score + confidence & \textbf{0.77} & \textbf{59} & $7.2\times10^{-39}$\\
\bottomrule
\end{tabular}
\caption{Correlation of judge score and confidence with GPT scores (\texttt{gpt-5.6sol} quality ratings) on 200 responses. The multiple correlation coefficient $R$ is the Pearson correlation between fitted and observed GPT scores; $R^2$ is in-sample explained variation. The $p$-values are from overall regression F-tests.}
\label{tab:judge-audit}
\end{table}

\paragraph{Higher judge scores and confidence are associated with higher GPT scores.}
We ask \texttt{gpt-5.6sol} \citep{openai2026gpt56} to rate 200 randomly sampled saved training responses, one per question, using the same 1--5 quality rubric as the training judge. These ratings, termed \emph{GPT scores}, serve only as an external measure of response quality and are not used to train \method{}. Table~\ref{tab:judge-audit} reports linear regressions that estimate GPT scores from judge score, judge confidence, or both. In the joint regression, higher judge scores are associated with higher GPT scores at the same confidence, and higher judge confidence is associated with higher GPT scores at the same judge score.

We report the multiple correlation coefficient $R$, defined as the Pearson correlation between fitted and observed GPT scores. Using judge score and confidence together gives $R=0.77$ and explains 59\% of the in-sample variation in GPT scores, compared with 21\% for judge score alone and 39\% for judge confidence alone. Adding confidence to score increases explained variation by $\Delta R^2=0.38$. Thus, judge confidence helps distinguish responses with the same judge score but different GPT scores.

These results support combining judge score and confidence to assess response quality. Section~\ref{sec:ablations} tests the product of normalized judge score and judge confidence as a training weight. GPT scores are external quality assessments, not verified correctness labels. Appendix~\ref{sec:implementation} gives the regression definitions and coefficients.

\subsection{Training Dynamics (RQ3)}
\label{sec:dynamics}

In this experiment, we analyze the Qwen3-4B run with $\alpha=0.5$. Figures~\ref{fig:dynamics} and~\ref{fig:student-groups} show changes in judge assessments and student uncertainty during training; Table~\ref{tab:dynamics-counts} in Appendix~\ref{sec:dynamics-counts} provides counts and detailed comparisons.

At reported steps 100, 200, and 300, scores 4 and 5 account for at least 99.2\% of completed responses, with score 5 becoming more frequent (Figure~\ref{fig:dynamics}a). Score-only weighting therefore assigns almost all responses either 0.75 or 1.0. Confidence distinguishes attempts with the same score; RQ2 links higher confidence to higher GPT scores at the same judge score.

\textbf{Both the number and percentage of responses with judge score 5 and high confidence increase.} From step 50 to 300, this group grows from 319 to 638 responses, or 26.5\% to 42.8\% of completed score-4/5 responses (Figure~\ref{fig:student-groups}a; Table~\ref{tab:dynamics-counts}). Within score 5, the high-confidence proportion ($c_i^{\mathrm J}\geq0.7$) rises from 49.0\% to 68.5\% (Figure~\ref{fig:dynamics}b). The score-4/high-confidence group also grows as a share of score-4/5 responses; its within-score proportion rises from 35.3\% to 61.6\%, with fluctuations.

\textbf{Mean student uncertainty falls 3.86-fold from step 50 to 300.} Across completed score-4/5 responses, mean $u_i^{\mathrm S}$ falls from 0.2649 to 0.0687, and the middle half of uncertainties also shifts downward (Figure~\ref{fig:dynamics}c). Student predictions thus become more concentrated.

\textbf{Within the score-5/high-confidence group, both the number and percentage with low student uncertainty increase overall.} Here, low uncertainty means below the mean of all completed score-4/5 responses in the same summary. The count rises from 156 of 319 responses (48.9\%) at step 50 to 463 of 638 (72.6\%) at step 300 (Figure~\ref{fig:student-groups}b; Table~\ref{tab:dynamics-counts}). The percentage peaks at 74.9\% at step 200. This threshold measures relative uncertainty because the mean changes across summaries.

\textbf{From step 100 onward, low student uncertainty is more frequent in high-confidence than low-confidence groups at both scores.} At score 5, the percentages are similar at step 50 but reach 72.6\% versus 39.8\% at step 300 (Table~\ref{tab:dynamics-counts}). Among high-confidence responses, the below-mean percentage is also higher for score 5 than score 4 from step 100 onward (Figure~\ref{fig:student-groups}b).

Together, rising judge confidence and falling student uncertainty are consistent with consolidation, but these summaries do not track individual attempts being learned. The two changes affect the response weight in opposite directions: higher judge confidence at a fixed score raises the reliability contribution, while lower student uncertainty reduces the uncertainty contribution.

Further, accuracy reaches 70.47\% at step 150, then falls despite declining student uncertainty and increasing high-confidence score-5 judgments (Figure~\ref{fig:training-steps}, Appendix~\ref{sec:training-duration}). Greater predictive certainty and judge confidence therefore do not establish continued improvement in reasoning. A selected heat-calculation case illustrates retained correctness with increasing certainty: both responses are correct and receive score 5, while judge confidence rises from 0.572 to 0.771 and student uncertainty falls from 0.1728 to 0.0388. Table~\ref{tab:case-study} gives the complete final responses; Appendix~\ref{sec:case-study} gives the calculations and selection details.

\subsection{Ablations and Hyperparameters (RQ4)}
\label{sec:ablations}

In this analysis, we study response weighting, loss type, and the weighting coefficient $\alpha$ on Qwen3-4B using the eleven-benchmark macroaverage. To assess what judge confidence and student uncertainty add beyond the judge score, our weighting ablations compare against score-only weighting rather than uniform weighting, which assigns the same loss weight to every attempt. Teacher context and forward KL remain fixed across the weighting variants. The loss ablations retain the full weighting rule and change the divergence. Table~\ref{tab:ablations} summarizes the ablations; Table~\ref{tab:weight-full} in Appendix~\ref{sec:extended-weighting} gives the per-benchmark comparison with score-only weighting.

\begin{table}[t]
\centering

\small
\begin{tabular}{@{}l r@{}}
\toprule
\textbf{Response weighting} & \textbf{Average}\\
\midrule
Score only: $w_i=\bar s_i$ & 68.44\\
Uncertainty only: $w_i=u_i^{\mathrm S}$ & 69.93\\
Reliability only: $w_i=r_i^{\mathrm J}$ & 70.12\\
\method{}: $w_i=0.5 r_i^{\mathrm J}+0.5 u_i^{\mathrm S}$ & \textbf{70.47}\\
\midrule
\textbf{Loss type} & \textbf{Average}\\
\midrule
Forward KL & \textbf{70.47}\\
Reverse KL & 69.85\\
Jensen--Shannon divergence & 69.12\\
\bottomrule
\end{tabular}
\caption{Response-weighting and loss ablations on Qwen3-4B, reporting average accuracy (Avg@4, \%) across eleven benchmarks. Weighting variants use the forward-KL configuration; loss variants use the full weighting rule at $\alpha=0.5$. Jensen--Shannon divergence uses mixture coefficient 0.5.}
\label{tab:ablations}
\end{table}

\noindent\emph{Ablation studies.}
\textbf{Adding judge confidence improves average accuracy by 1.68.} Score-only weighting, $w_i=\bar{s}_i$, reaches 68.44\%; multiplying the normalized score by judge confidence, which gives reliability-only weighting $w_i=r_i^{\mathrm J}=\bar{s}_i c_i^{\mathrm J}$, raises accuracy to 70.12\%. Score-only weighting assigns the same weight to all attempts with the same rating; confidence differentiates attempts within each score level. The gain is consistent with the quality audit, in which confidence explains additional variation in external ratings.

\textbf{Combining reliability and uncertainty gives the highest average accuracy among all the weighting variants.} Reliability-only weighting prioritizes assessed quality, while uncertainty-only weighting prioritizes diffuse student predictions. The full rule combines both priorities and reaches 70.47\%, exceeding reliability-only weighting by 0.35 and uncertainty-only weighting by 0.54. It is also higher than score-only weighting on all eleven benchmarks, with a macroaverage gain of 2.03. Adding confidence gives the larger observed gain (1.68), while adding student uncertainty to reliability gives a smaller further gain in this single-run comparison. Both single-signal variants exceed the strongest baseline average, while their combination gives a further improvement.

\textbf{Forward KL gives the highest average among the tested loss types.} Response weights determine each attempt's contribution, while the loss type determines how teacher and student token distributions are matched. With the full weighting rule and $\alpha=0.5$, the forward-KL configuration reaches 70.47\%, compared with 69.85\% for reverse KL and 69.12\% for Jensen--Shannon divergence with mixture coefficient 0.5. 


\begin{figure}[t]
\centering
\includegraphics[width=0.75\columnwidth]{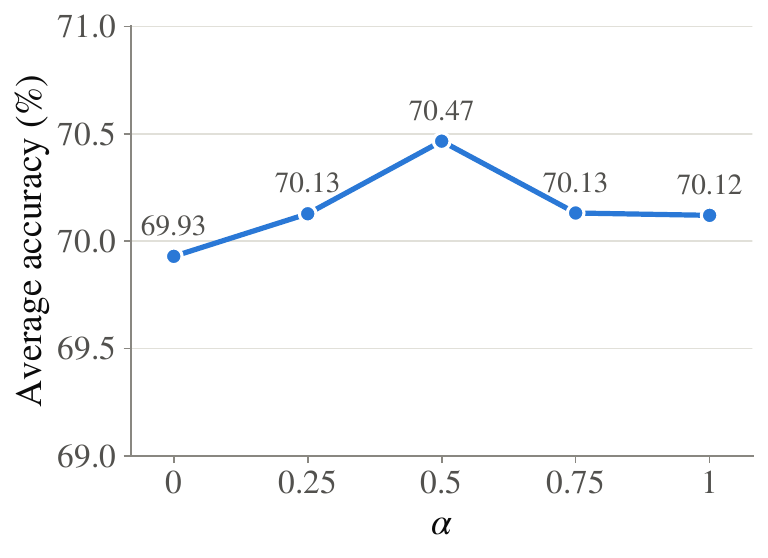}
\caption{Effect of the mixing coefficient $\alpha$ on Qwen3-4B \method{}, measured by average accuracy across eleven benchmarks. All settings use the dynamic teacher. The endpoints, $\alpha=0$ (uncertainty only) and $\alpha=1$ (reliability only), match the corresponding runs in Table~\ref{tab:ablations}.}
\label{fig:alpha}
\end{figure}

\noindent\emph{Hyperparameter study.}
\textbf{$\alpha=0.5$ gives the highest observed average accuracy.} In Equation~\ref{eq:weight}, $\alpha=0$ gives uncertainty-only weighting and $\alpha=1$ gives reliability-only weighting, matching the corresponding ablations in Table~\ref{tab:ablations}. Larger $\alpha$ assigns a higher coefficient to judged reliability, emphasizing quality supported by judge confidence; smaller $\alpha$ assigns a higher coefficient to remaining student uncertainty. Figure~\ref{fig:alpha} reports 69.93\%, 70.13\%, 70.47\%, 70.13\%, and 70.12\% for $\alpha=0$, 0.25, 0.5, 0.75, and 1, respectively. The full range is 0.54, and all three intermediate settings reach at least 70.13\%. The highest observed average favors using both signals in this configuration.

\section{Conclusion}

In this work, we address two challenges in self-distillation: obtaining teacher guidance without supplied reference solutions, and assessing self-generated attempts without reliable correctness feedback. \method{} uses the attempts as teacher context and weights distillation by judged reliability and student uncertainty. Across eleven benchmarks, it improves Qwen3-4B and Qwen3-8B macroaverage accuracy by 8.90 and 2.10 over the base models, respectively. At 4B, adding judge confidence improves average accuracy, and combining reliability with uncertainty gives the highest observed average among the weighting variants. External ratings support judge confidence as a quality signal, while training summaries show more high-confidence judgments alongside lower student uncertainty.

\clearpage
\section*{Limitations}

\paragraph{Model coverage.}
Our experiments use Qwen3-4B and Qwen3-8B. Further evaluation across model families and sizes is needed to establish how well \method{} generalizes beyond these backbones.

\paragraph{Judge design.}
For efficiency, the judge assesses completed reasoning responses with thinking mode disabled, directly reading rating-token logits without generating an evaluation rationale. Whether enabling judge thinking or updating the judge improves assessment quality remains untested. Enabling thinking would add decoding cost and latency, depending on the reasoning budget; judge-update costs depend on whether updates reuse student parameters or require separate optimization. We also do not compare larger judges or judges from other model families within \method{}, so their effects on accuracy and cost remain untested.

\paragraph{Self-judgment and learning dynamics.}
The judge can confidently endorse incorrect responses and share biases with the generating model. Its confidence measures concentration over ratings, rather than a calibrated probability of correctness. The additive rule also assigns uncertainty-based weight when judged reliability is low. The 200-response audit measures in-sample agreement with automated quality ratings, rather than verified correctness. Training summaries follow changing response populations and describe associations rather than causal learning mechanisms. The later decline in accuracy despite falling entropy shows that greater predictive certainty does not guarantee better reasoning.

\paragraph{Component evidence.}
Component studies use Qwen3-4B. Judged reliability and student uncertainty both lie in $[0,1]$; their interaction and relative influence on learning across training stages warrant deeper analysis.

\paragraph{Cost and scope.}
Each attempt requires a teacher forward pass conditioned on the complete attempt and a separate judge forward pass. The teacher does not generate a second response. We do not report total compute across all experiments or compare methods at matched compute budgets, and we do not analyze response length or truncation rates. Experiments use one NaturalReasoning training selection, English questions, and thinking mode for response generation. All benchmarks have automatically checkable answers, so performance on open-ended tasks remains untested.

\section*{Ethical Considerations}
\label{sec:ethics}

This study uses publicly available reasoning resources and model-generated responses. NaturalReasoning is released under CC BY-NC 4.0, and the Qwen3-4B and Qwen3-8B backbones under Apache 2.0.\footnote{See the official \href{https://huggingface.co/datasets/facebook/natural_reasoning}{NaturalReasoning dataset card} and \href{https://huggingface.co/Qwen/Qwen3-4B}{Qwen3-4B}/\href{https://huggingface.co/Qwen/Qwen3-8B}{Qwen3-8B model cards}.} The dataset's Llama3.3-70B-Instruct responses, which only the SFT and OPSD baselines use, carry additional obligations under the \href{https://huggingface.co/meta-llama/Llama-3.3-70B-Instruct/blob/main/LICENSE}{Llama 3.3 Community License} when used to train distributed models. Reuse and redistribution should respect these licenses and the terms of individual evaluation resources, including the training dataset's noncommercial restriction.

Self-distillation can reinforce factual errors and social biases inherited from the model or its training data. The teacher and judge share a pretrained backbone, so errors may persist even when judge confidence increases and student uncertainty decreases. The proposed weighting signals are fallible assessments, so more confident responses should not be treated as verified answers.

The external quality check uses automated \texttt{gpt-5.6sol} ratings of 200 responses. It provides evidence about assessed response quality, rather than human verification or a safety evaluation. Our experiments do not measure fairness, privacy, or robustness to harmful requests. Deployment, especially in high-stakes settings, would require separate assessment and safeguards appropriate to the intended use.

\bibliography{custom,anthology-1,anthology-2}

@misc{hinton2015distillingknowledgeneuralnetwork,
 title={Distilling the Knowledge in a Neural Network},
 author={Geoffrey Hinton and Oriol Vinyals and Jeff Dean},
 year={2015}, eprint={1503.02531}, archivePrefix={arXiv}, primaryClass={stat.ML},
 url={https://arxiv.org/abs/1503.02531}
}

@inproceedings{zelikman2022starbootstrappingreasoningreasoning,
 title={{STaR}: Bootstrapping Reasoning With Reasoning},
 author={Eric Zelikman and Yuhuai Wu and Jesse Mu and Noah Goodman},
 booktitle={Advances in Neural Information Processing Systems},
 editor={S. Koyejo and S. Mohamed and A. Agarwal and D. Belgrave and K. Cho and A. Oh},
 publisher={Curran Associates, Inc.}, volume={35}, pages={15476--15488}, year={2022},
 doi={10.52202/068431-1126},
 url={https://proceedings.neurips.cc/paper_files/paper/2022/file/639a9a172c044fbb64175b5fad42e9a5-Paper-Conference.pdf}
}

@inproceedings{madaan2023selfrefineiterativerefinementselffeedback,
 title={{Self-Refine}: Iterative Refinement with Self-Feedback},
 author={Aman Madaan and Niket Tandon and Prakhar Gupta and Skyler Hallinan and Luyu Gao and Sarah Wiegreffe and Uri Alon and Nouha Dziri and Shrimai Prabhumoye and Yiming Yang and Shashank Gupta and Bodhisattwa Prasad Majumder and Katherine Hermann and Sean Welleck and Amir Yazdanbakhsh and Peter Clark},
 booktitle={Advances in Neural Information Processing Systems},
 editor={A. Oh and T. Naumann and A. Globerson and K. Saenko and M. Hardt and S. Levine},
 publisher={Curran Associates, Inc.}, volume={36}, pages={46534--46594}, year={2023},
 doi={10.52202/075280-2019},
 url={https://proceedings.neurips.cc/paper_files/paper/2023/file/91edff07232fb1b55a505a9e9f6c0ff3-Paper-Conference.pdf}
}

@inproceedings{zhao2026learningreasonexternalrewards,
 title={Learning to Reason without External Rewards},
 author={Xuandong Zhao and Zhewei Kang and Aosong Feng and Sergey Levine and Dawn Song},
 booktitle={The Fourteenth International Conference on Learning Representations},
 year={2026}, url={https://arxiv.org/abs/2505.19590}
}

@inproceedings{shenfeld2026selfdistillationenablescontinuallearning,
 title={Self-Distillation Enables Continual Learning},
 author={Idan Shenfeld and Mehul Damani and Jonas H{\"u}botter and Pulkit Agrawal},
 booktitle={Proceedings of the 43rd International Conference on Machine Learning},
 pages={111542--111558}, volume={306},
 series={Proceedings of Machine Learning Research}, publisher={PMLR},
 year={2026}, url={https://proceedings.mlr.press/v306/shenfeld26a.html}
}

@inproceedings{zhao2026selfdistilledreasoneronpolicyselfdistillation,
 title={Self-Distilled Reasoner: On-Policy Self-Distillation for Large Language Models},
 author={Siyan Zhao and Zhihui Xie and Mengchen Liu and Jing Huang and Guan Pang and Feiyu Chen and Aditya Grover},
 booktitle={Proceedings of the 43rd International Conference on Machine Learning},
 pages={162433--162448}, volume={306},
 series={Proceedings of Machine Learning Research}, publisher={PMLR},
 year={2026}, url={https://proceedings.mlr.press/v306/zhao26be.html}
}

@inproceedings{gu2026rethinkingrewardsupervisionrubricconditioned,
 title={Rethinking Reward Supervision: Rubric-Conditioned Self-Distillation},
 author={Siyi Gu and Jialin Chen and Sophia Zhou and Arman Cohan and Rex Ying},
 booktitle={Third Conference on Language Modeling},
 year={2026}, url={https://arxiv.org/abs/2606.19327}
}

@misc{pei2026negativeselfdistillationlearningreason,
 title={Negative Self-Distillation: Learning to Reason by Avoiding Flaws},
 author={Rongcan Pei and Zhepei Wei and Shuyao Xu and Xinyu Zhu and Wei-Lin Chen and Yu Meng},
 year={2026}, eprint={2609.11699}, archivePrefix={arXiv}, primaryClass={cs.CL},
 url={https://arxiv.org/abs/2609.11699}
}

@misc{rezaei2026rubricguidedselfdistillationposttrainingrubric,
 title={Rubric-Guided Self-Distillation: Post-Training Without Rubric Verifiers},
 author={MohammadHossein Rezaei and Anas Mahmoud and Zihao Wang and Utkarsh Tyagi and Advait Gosai and Razvan-Gabriel Dumitru and Aakash Sabharwal and Bing Liu and Yunzhong He},
 year={2026}, eprint={2606.12507}, archivePrefix={arXiv}, primaryClass={cs.LG},
 url={https://arxiv.org/abs/2606.12507}
}

@misc{ye2026llmasacoachexperientiallearningnonverifiable,
 title={{LLM}-as-a-Coach: Experiential Learning for Non-Verifiable Tasks},
 author={Tianzhu Ye and Li Dong and Guanheng Chen and He Zhu and Xun Wu and Shaohan Huang and Furu Wei},
 year={2026}, eprint={2607.18110}, archivePrefix={arXiv}, primaryClass={cs.LG},
 url={https://arxiv.org/abs/2607.18110}
}

@inproceedings{wei2026confidenceorchestratedselfevolutionuncertainllm,
 title={{COSE}: Confidence-Orchestrated Self-Evolution for Effective {LLM} Reasoning},
 author={Bowen Wei and Nan Wang and Yuqing Zhou and Jinhao Pan and Ziwei Zhu},
 booktitle={Proceedings of the 2026 Conference on Empirical Methods in Natural Language Processing},
 publisher={Association for Computational Linguistics},
 year={2026},
 note={To appear, main conference},
 url={https://arxiv.org/abs/2605.28010}
}

@inproceedings{agarwal2024onpolicydistillationlanguagemodels,
 title={On-Policy Distillation of Language Models: Learning from Self-Generated Mistakes},
 author={Rishabh Agarwal and Nino Vieillard and Yongchao Zhou and Piotr Stanczyk and Sabela Ramos and Matthieu Geist and Olivier Bachem},
 booktitle={The Twelfth International Conference on Learning Representations},
 year={2024}, url={https://arxiv.org/abs/2306.13649}
}

@inproceedings{yuan2024selfrewardinglanguagemodels,
 title={Self-Rewarding Language Models},
 author={Weizhe Yuan and Richard Yuanzhe Pang and Kyunghyun Cho and Xian Li and Sainbayar Sukhbaatar and Jing Xu and Jason E Weston},
 booktitle={Proceedings of the 41st International Conference on Machine Learning},
 year={2024}, pages={57905--57923}, volume={235},
 series={Proceedings of Machine Learning Research}, publisher={PMLR},
 url={https://proceedings.mlr.press/v235/yuan24d.html}
}

@inproceedings{chen2023softmatchaddressingquantityqualitytradeoff,
 title={{SoftMatch}: Addressing the Quantity-Quality Trade-off in Semi-supervised Learning},
 author={Hao Chen and Ran Tao and Yue Fan and Yidong Wang and Jindong Wang and Bernt Schiele and Xing Xie and Bhiksha Raj and Marios Savvides},
 booktitle={The Eleventh International Conference on Learning Representations},
 year={2023}, url={https://arxiv.org/abs/2301.10921}
}

@inproceedings{guo2017calibrationmodernneuralnetworks,
 title={On Calibration of Modern Neural Networks},
 author={Chuan Guo and Geoff Pleiss and Yu Sun and Kilian Q. Weinberger},
 booktitle={Proceedings of the 34th International Conference on Machine Learning},
 series={Proceedings of Machine Learning Research}, volume={70}, pages={1321--1330},
 publisher={PMLR}, year={2017},
 url={https://proceedings.mlr.press/v70/guo17a.html}
}

@inproceedings{hu2021loralowrankadaptationlarge,
 title={{LoRA}: Low-Rank Adaptation of Large Language Models},
 author={Edward J. Hu and Yelong Shen and Phillip Wallis and Zeyuan Allen-Zhu and Yuanzhi Li and Shean Wang and Lu Wang and Weizhu Chen},
 booktitle={International Conference on Learning Representations},
 year={2022}, url={https://arxiv.org/abs/2106.09685}
}

@inproceedings{yuan2025naturalreasoningreasoningwild28m,
 title={{NaturalReasoning}: Reasoning in the Wild with {2.8M} Challenging Questions},
 author={Weizhe Yuan and Jane Yu and Song Jiang and Karthik Padthe and Yang Li and Dong Wang and Ilia Kulikov and Kyunghyun Cho and Yuandong Tian and Jason Weston and Xian Li},
 booktitle={Advances in Neural Information Processing Systems},
 editor={D. Belgrave and C. Zhang and H. Lin and R. Pascanu and P. Koniusz and M. Ghassemi and N. Chen},
 publisher={Curran Associates, Inc.}, volume={38}, year={2025},
 note={Datasets and Benchmarks Track}, doi={10.52202/085713-2246},
 url={https://proceedings.neurips.cc/paper_files/paper/2025/file/60b22138917015fcba7165670e6f453c-Paper-Datasets_and_Benchmarks_Track.pdf}
}

@inproceedings{ouyang2022traininglanguagemodelsfollow,
  title={Training language models to follow instructions with human feedback},
  author={Long Ouyang and Jeffrey Wu and Xu Jiang and Diogo Almeida and Carroll Wainwright and Pamela Mishkin and Chong Zhang and Sandhini Agarwal and Katarina Slama and Alex Ray and John Schulman and Jacob Hilton and Fraser Kelton and Luke Miller and Maddie Simens and Amanda Askell and Peter Welinder and Paul F Christiano and Jan Leike and Ryan Lowe},
  booktitle={Advances in Neural Information Processing Systems},
  editor={S. Koyejo and S. Mohamed and A. Agarwal and D. Belgrave and K. Cho and A. Oh},
  publisher={Curran Associates, Inc.}, volume={35}, pages={27730--27744}, year={2022},
  doi={10.52202/068431-2011},
  url={https://proceedings.neurips.cc/paper_files/paper/2022/file/b1efde53be364a73914f58805a001731-Paper-Conference.pdf}
}

@misc{deepseekai2025deepseekr1,
  title={{DeepSeek-R1}: Incentivizing Reasoning Capability in {LLMs} via Reinforcement Learning},
  author={{DeepSeek-AI} and Daya Guo and Dejian Yang and Haowei Zhang and Junxiao Song and others},
  year={2025}, eprint={2501.12948}, archivePrefix={arXiv}, primaryClass={cs.CL},
  url={https://arxiv.org/abs/2501.12948}
}

@misc{askell2021generallanguageassistantlaboratory,
  title={A General Language Assistant as a Laboratory for Alignment},
  author={Amanda Askell and Yuntao Bai and Anna Chen and Dawn Drain and Deep Ganguli and Tom Henighan and Andy Jones and Nicholas Joseph and Ben Mann and Nova DasSarma and Nelson Elhage and Zac Hatfield-Dodds and Danny Hernandez and Jackson Kernion and Kamal Ndousse and Catherine Olsson and Dario Amodei and Tom Brown and Jack Clark and Sam McCandlish and Chris Olah and Jared Kaplan},
  year={2021}, eprint={2112.00861}, archivePrefix={arXiv}, primaryClass={cs.CL},
  url={https://arxiv.org/abs/2112.00861}
}

@misc{snell2022learningdistillingcontext,
  title={Learning by Distilling Context},
  author={Charlie Snell and Dan Klein and Ruiqi Zhong},
  year={2022}, eprint={2209.15189}, archivePrefix={arXiv}, primaryClass={cs.CL},
  url={https://arxiv.org/abs/2209.15189}
}

@misc{ye2026onpolicycontextdistillationlanguage,
  title={On-Policy Context Distillation for Language Models},
  author={Tianzhu Ye and Li Dong and Xun Wu and Shaohan Huang and Furu Wei},
  year={2026}, eprint={2602.12275}, archivePrefix={arXiv}, primaryClass={cs.CL},
  url={https://arxiv.org/abs/2602.12275}
}

@article{vapnik2009newlearningparadigmlearning,
  title={A new learning paradigm: Learning using privileged information},
  author={Vladimir Vapnik and Akshay Vashist},
  journal={Neural Networks},
  volume={22}, number={5-6}, pages={544--557}, year={2009},
  doi={10.1016/j.neunet.2009.06.042}
}

@inproceedings{lopezpaz2016unifyingdistillationprivilegedinformation,
  title={Unifying distillation and privileged information},
  author={David Lopez-Paz and L{\'e}on Bottou and Bernhard Sch{\"o}lkopf and Vladimir Vapnik},
  booktitle={International Conference on Learning Representations},
  year={2016}, url={https://arxiv.org/abs/1511.03643}
}

@misc{ichihara2026privilegedsolutionscontextinducedteacher,
  title={Privileged Solutions or Context-Induced Teacher Behavior? Dissecting On-Policy Self-Distillation},
  author={Yuki Ichihara and Naoto Iwase and Mohammad Atif Quamar and Junpei Komiyama},
  year={2026}, eprint={2608.09228}, archivePrefix={arXiv}, primaryClass={cs.LG},
  url={https://arxiv.org/abs/2608.09228}
}

@misc{shrestha2026rethinkingprivilegedinformationonpolicy,
  title={Rethinking Privileged Information in On-Policy Self-Distillation},
  author={Samyak Shrestha and Alexander Tessier},
  year={2026}, eprint={2608.18271}, archivePrefix={arXiv}, primaryClass={cs.LG},
  url={https://arxiv.org/abs/2608.18271}
}

@inproceedings{hubotter2026reinforcementlearningselfdistillation,
  title={Reinforcement Learning via Self-Distillation},
  author={Jonas H{\"u}botter and Frederike L{\"u}beck and Lejs Deen Behric and Anton Baumann and Marco Bagatella and Daniel Marta and Ido Hakimi and Idan Shenfeld and Kleine Buening, Thomas and Carlos Guestrin and Andreas Krause},
  booktitle={Proceedings of the 43rd International Conference on Machine Learning},
  pages={48677--48721}, volume={306},
  series={Proceedings of Machine Learning Research}, publisher={PMLR},
  year={2026}, url={https://proceedings.mlr.press/v306/hubotter26a.html}
}

@misc{he2026selfdistillationzeroselfrevisionturns,
  title={Self-Distillation Zero: Self-Revision Turns Binary Rewards into Dense Supervision},
  author={Yinghui He and Simran Kaur and Adithya Bhaskar and Yongjin Yang and Jiarui Liu and Narutatsu Ri and Liam Fowl and Abhishek Panigrahi and Danqi Chen and Sanjeev Arora},
  year={2026}, eprint={2604.12002}, archivePrefix={arXiv}, primaryClass={cs.CL},
  url={https://arxiv.org/abs/2604.12002}
}

@misc{cai2026h2sdhybridhindsightselfdistillation,
  title={{H$^2$SD}: Hybrid Hindsight Self-Distillation},
  author={Qiye Cai and Yichuan Ma and Peiji Li and Yongkang Chen and Qipeng Guo and Yicheng Zou and Linyang Li and Xiaocheng Feng and Bing Qin},
  year={2026}, eprint={2607.18955}, archivePrefix={arXiv}, primaryClass={cs.LG},
  url={https://arxiv.org/abs/2607.18955}
}

@misc{hossain2026overcomingscalinglimitsonpolicy,
  title={Overcoming Scaling Limits in On-Policy Self-Distillation for {LLM} Reasoning},
  author={Md. Ismail Hossain and Humaira Kousar and Isidora Chara Tourni},
  year={2026}, eprint={2609.37915}, archivePrefix={arXiv}, primaryClass={cs.LG},
  url={https://arxiv.org/abs/2609.37915}
}

@misc{feng2026pastprivilegedadaptationcomplete,
  title={{PAST}: Privileged Adaptation from Complete Student Trajectories for On-Policy Self-Distillation},
  author={Yangyang Feng and Zhuoyan Feng and Junlan Chen},
  year={2026}, eprint={2608.08726}, archivePrefix={arXiv}, primaryClass={cs.LG},
  url={https://arxiv.org/abs/2608.08726}
}

@misc{gkountouras2026consensusprivilegedcontextlabelfree,
  title={Consensus as Privileged Context for Label-Free Self-Distillation},
  author={John Gkountouras and Josip Juki{\'c} and Ivan Titov},
  year={2026}, eprint={2607.13643}, archivePrefix={arXiv}, primaryClass={cs.LG},
  url={https://arxiv.org/abs/2607.13643}
}

@misc{li2026onpolicyselfdistillationanysupervision,
  title={On-Policy Self-Distillation without Any Supervision},
  author={Yijiang Li and Bingyang Wang and Yijun Liang and Yunjie Tian and Di Fu and Nuno Vasconcelos},
  year={2026}, eprint={2608.06296}, archivePrefix={arXiv}, primaryClass={cs.LG},
  url={https://arxiv.org/abs/2608.06296}
}

@misc{guo2026learningconsensusdisagreementunsupervised,
  title={Learning from Consensus and Disagreement: Unsupervised On-Policy Self-Distillation with Minority-Trajectory Contrast},
  author={Jiaxin Guo and Yanwei Yue and Xuanbo Fan and Chunyu Yang and Yan Zhang},
  year={2026}, eprint={2608.08764}, archivePrefix={arXiv}, primaryClass={cs.LG},
  url={https://arxiv.org/abs/2608.08764}
}

@misc{gulcehre2023reinforcedselftrainingrestlanguage,
  title={Reinforced Self-Training ({ReST}) for Language Modeling},
  author={Caglar Gulcehre and Le Paine, Tom and Srivatsan Srinivasan and Ksenia Konyushkova and Lotte Weerts and Abhishek Sharma and Aditya Siddhant and Alex Ahern and Miaosen Wang and Chenjie Gu and Wolfgang Macherey and Arnaud Doucet and Orhan Firat and Nando de Freitas},
  year={2023}, eprint={2308.08998}, archivePrefix={arXiv}, primaryClass={cs.CL},
  url={https://arxiv.org/abs/2308.08998}
}

@article{dong2023raftrewardrankedfinetuning,
  title={{RAFT}: Reward rAnked FineTuning for Generative Foundation Model Alignment},
  author={Hanze Dong and Wei Xiong and Deepanshu Goyal and Yihan Zhang and Winnie Chow and Rui Pan and Shizhe Diao and Jipeng Zhang and KaShun Shum and Tong Zhang},
  journal={Transactions on Machine Learning Research},
  issn={2835-8856}, year={2023},
  url={https://openreview.net/forum?id=m7p5O7zblY}
}

@article{singh2024beyondhumandatascaling,
  title={Beyond Human Data: Scaling Self-Training for Problem-Solving with Language Models},
  author={Avi Singh and John D Co-Reyes and Rishabh Agarwal and Ankesh Anand and Piyush Patil and Xavier Garcia and Peter J Liu and James Harrison and Jaehoon Lee and Kelvin Xu and Aaron T Parisi and Abhishek Kumar and Alexander A Alemi and Alex Rizkowsky and Azade Nova and Ben Adlam and Bernd Bohnet and Gamaleldin Fathy Elsayed and Hanie Sedghi and Igor Mordatch and Isabelle Simpson and Izzeddin Gur and Jasper Snoek and Jeffrey Pennington and Jiri Hron and Kathleen Kenealy and Kevin Swersky and Kshiteej Mahajan and Laura A Culp and Lechao Xiao and Maxwell Bileschi and Noah Constant and Roman Novak and Rosanne Liu and Tris Warkentin and Yundi Qian and Yamini Bansal and Ethan Dyer and Behnam Neyshabur and Jascha Sohl-Dickstein and Noah Fiedel},
  journal={Transactions on Machine Learning Research},
  issn={2835-8856}, year={2024}, note={Expert Certification},
  url={https://openreview.net/forum?id=lNAyUngGFK}
}

@inproceedings{agarwal2025unreasonableeffectivenessentropyminimization,
  title={The Unreasonable Effectiveness of Entropy Minimization in {LLM} Reasoning},
  author={Shivam Agarwal and Zimin Zhang and Lifan Yuan and Jiawei Han and Hao Peng},
  booktitle={Advances in Neural Information Processing Systems},
  editor={D. Belgrave and C. Zhang and H. Lin and R. Pascanu and P. Koniusz and M. Ghassemi and N. Chen},
  publisher={Curran Associates, Inc.}, volume={38}, pages={107150--107180}, year={2025},
  doi={10.52202/085713-3573},
  url={https://proceedings.neurips.cc/paper_files/paper/2025/file/9a14cf2ee36cf2524f54d8f196a6734c-Paper-Conference.pdf}
}

@misc{prabhudesai2025maximizingconfidencealoneimproves,
  title={Maximizing Confidence Alone Improves Reasoning},
  author={Mihir Prabhudesai and Lili Chen and Alex Ippoliti and Katerina Fragkiadaki and Hao Liu and Deepak Pathak},
  year={2025}, eprint={2505.22660}, archivePrefix={arXiv}, primaryClass={cs.LG},
  url={https://arxiv.org/abs/2505.22660}
}

@misc{nicolicioiu2026onpolicyselfdistillationsampleddemonstrations,
  title={On-Policy Self-Distillation with Sampled Demonstrations Reduces Output Diversity},
  author={Andrei Liviu Nicolicioiu and Mohammad Pezeshki and Aaron Courville},
  year={2026}, eprint={2606.26091}, archivePrefix={arXiv}, primaryClass={cs.LG},
  url={https://arxiv.org/abs/2606.26091}
}

@inproceedings{mukherjee2020uncertaintyawareselftrainingfewshot,
  title={Uncertainty-aware Self-training for Few-shot Text Classification},
  author={Subhabrata Mukherjee and Ahmed Awadallah},
  booktitle={Advances in Neural Information Processing Systems},
  editor={H. Larochelle and M. Ranzato and R. Hadsell and M.F. Balcan and H. Lin},
  publisher={Curran Associates, Inc.}, volume={33}, pages={21199--21212}, year={2020},
  url={https://proceedings.neurips.cc/paper_files/paper/2020/file/f23d125da1e29e34c552f448610ff25f-Paper.pdf}
}

@misc{ke2026respectingselfuncertaintyonpolicyselfdistillation,
  title={Respecting Self-Uncertainty in On-Policy Self-Distillation for Efficient {LLM} Reasoning},
  author={Junlong Ke and Zichen Wen and Weijia Li and Conghui He and Linfeng Zhang},
  year={2026}, eprint={2605.13255}, archivePrefix={arXiv}, primaryClass={cs.AI},
  url={https://arxiv.org/abs/2605.13255}
}

@inproceedings{zheng2023judgingllmasajudgemtbench,
  title={Judging {LLM}-as-a-Judge with {MT-Bench} and {Chatbot Arena}},
  author={Lianmin Zheng and Wei-Lin Chiang and Ying Sheng and Siyuan Zhuang and Zhanghao Wu and Yonghao Zhuang and Zi Lin and Zhuohan Li and Dacheng Li and Eric Xing and Hao Zhang and Joseph Gonzalez and Ion Stoica},
  booktitle={Advances in Neural Information Processing Systems},
  editor={A. Oh and T. Naumann and A. Globerson and K. Saenko and M. Hardt and S. Levine},
  publisher={Curran Associates, Inc.}, volume={36}, pages={46595--46623}, year={2023},
  note={Datasets and Benchmarks Track}, doi={10.52202/075280-2020},
  url={https://proceedings.neurips.cc/paper_files/paper/2023/file/91f18a1287b398d378ef22505bf41832-Paper-Datasets_and_Benchmarks.pdf}
}

@inproceedings{panickssery2024llmevaluatorsrecognizefavor,
  title={{LLM} Evaluators Recognize and Favor Their Own Generations},
  author={Arjun Panickssery and Samuel R. Bowman and Shi Feng},
  booktitle={Advances in Neural Information Processing Systems},
  editor={A. Globerson and L. Mackey and D. Belgrave and A. Fan and U. Paquet and J. Tomczak and C. Zhang},
  publisher={Curran Associates, Inc.}, volume={37}, pages={68772--68802}, year={2024},
  doi={10.52202/079017-2197},
  url={https://proceedings.neurips.cc/paper_files/paper/2024/file/7f1f0218e45f5414c79c0679633e47bc-Paper-Conference.pdf}
}

@misc{huang2026whenclippingreversescorrection,
  title={When Clipping Reverses Correction: Failure Dynamics of Pointwise Forward-{KL} On-Policy Self-Distillation},
  author={Di Huang and Hao Li and Yixin Chen and Fuhai Li},
  year={2026}, eprint={2609.38995}, archivePrefix={arXiv}, primaryClass={cs.CL},
  url={https://arxiv.org/abs/2609.38995}
}

@inproceedings{rein2024gpqa,
  title={{GPQA}: A Graduate-Level {Google}-Proof {Q\&A} Benchmark},
  author={David Rein and Betty Li Hou and Asa Cooper Stickland and Jackson Petty and Richard Yuanzhe Pang and Julien Dirani and Julian Michael and Samuel R. Bowman},
  booktitle={First Conference on Language Modeling},
  year={2024},
  url={https://openreview.net/forum?id=Ti67584b98}
}

@inproceedings{wang2024scibench,
  title={{SciBench}: Evaluating College-Level Scientific Problem-Solving Abilities of Large Language Models},
  author={Xiaoxuan Wang and Ziniu Hu and Pan Lu and Yanqiao Zhu and Jieyu Zhang and Satyen Subramaniam and Arjun R. Loomba and Shichang Zhang and Yizhou Sun and Wei Wang},
  booktitle={Proceedings of the 41st International Conference on Machine Learning},
  volume={235}, series={Proceedings of Machine Learning Research},
  pages={50622--50649}, publisher={PMLR}, year={2024},
  url={https://proceedings.mlr.press/v235/wang24z.html}
}

@inproceedings{hendrycks2021measuring,
  title={Measuring Massive Multitask Language Understanding},
  author={Dan Hendrycks and Collin Burns and Steven Basart and Andy Zou and Mantas Mazeika and Dawn Song and Jacob Steinhardt},
  booktitle={International Conference on Learning Representations},
  year={2021},
  url={https://openreview.net/forum?id=d7KBjmI3GmQ}
}

@inproceedings{white2025livebench,
  title={{LiveBench}: A Challenging, Contamination-Limited {LLM} Benchmark},
  author={Colin White and Samuel Dooley and Manley Roberts and Arka Pal and Benjamin Feuer and Siddhartha Jain and Ravid Shwartz-Ziv and Neel Jain and Khalid Saifullah and Sreemanti Dey and Shubh-Agrawal and Sandeep Singh Sandha and Siddartha Venkat Naidu and Chinmay Hegde and Yann LeCun and Tom Goldstein and Willie Neiswanger and Micah Goldblum},
  booktitle={The Thirteenth International Conference on Learning Representations},
  year={2025},
  url={https://openreview.net/forum?id=sKYHBTAxVa}
}

@misc{cobbe2021trainingverifiers,
  title={Training Verifiers to Solve Math Word Problems},
  author={Karl Cobbe and Vineet Kosaraju and Mohammad Bavarian and Mark Chen and Heewoo Jun and Lukasz Kaiser and Matthias Plappert and Jerry Tworek and Jacob Hilton and Reiichiro Nakano and Christopher Hesse and John Schulman},
  year={2021}, eprint={2110.14168}, archivePrefix={arXiv}, primaryClass={cs.LG},
  url={https://arxiv.org/abs/2110.14168}
}

@inproceedings{hendrycks2021measuringmathematical,
  title={Measuring Mathematical Problem Solving With the {MATH} Dataset},
  author={Dan Hendrycks and Collin Burns and Saurav Kadavath and Akul Arora and Steven Basart and Eric Tang and Dawn Song and Jacob Steinhardt},
  booktitle={Proceedings of the Neural Information Processing Systems Track on Datasets and Benchmarks},
  volume={1}, year={2021},
  url={https://datasets-benchmarks-proceedings.neurips.cc/paper_files/paper/2021/hash/be83ab3ecd0db773eb2dc1b0a17836a1-Abstract-round2.html}
}

@misc{aimo2024amc,
  title={{AIMO Validation AMC}},
  author={{AI-MO}},
  year={2024},
  howpublished={Hugging Face dataset},
  url={https://huggingface.co/datasets/AI-MO/aimo-validation-amc}
}

@inproceedings{lewkowycz2022solving,
  title={Solving Quantitative Reasoning Problems with Language Models},
  author={Aitor Lewkowycz and Anders Andreassen and David Dohan and Ethan Dyer and Henryk Michalewski and Vinay Ramasesh and Ambrose Slone and Cem Anil and Imanol Schlag and Theo Gutman-Solo and Yuhuai Wu and Behnam Neyshabur and Guy Gur-Ari and Vedant Misra},
  booktitle={Advances in Neural Information Processing Systems},
  volume={35}, year={2022},
  doi={10.52202/068431-0278},
  url={https://proceedings.neurips.cc/paper_files/paper/2022/hash/18abbeef8cfe9203fdf9053c9c4fe191-Abstract.html}
}

@inproceedings{lin2025zebralogic,
  title={{ZebraLogic}: On the Scaling Limits of {LLMs} for Logical Reasoning},
  author={Bill Yuchen Lin and Le Bras, Ronan and Kyle Richardson and Ashish Sabharwal and Radha Poovendran and Peter Clark and Yejin Choi},
  booktitle={Proceedings of the 42nd International Conference on Machine Learning},
  volume={267}, series={Proceedings of Machine Learning Research},
  pages={37889--37905}, publisher={PMLR}, year={2025},
  url={https://proceedings.mlr.press/v267/lin25i.html}
}

@misc{qwen2025qwen3,
  title={{Qwen3} Technical Report},
  author={An Yang and Anfeng Li and Baosong Yang and Beichen Zhang and Binyuan Hui and Bo Zheng and Bowen Yu and Chang Gao and Chengen Huang and Chenxu Lv and Chujie Zheng and Dayiheng Liu and Fan Zhou and Fei Huang and Feng Hu and Hao Ge and Haoran Wei and Huan Lin and Jialong Tang and Jian Yang and Jianhong Tu and Jianwei Zhang and Jianxin Yang and Jiaxi Yang and Jing Zhou and Jingren Zhou and Junyang Lin and Kai Dang and Keqin Bao and Kexin Yang and Le Yu and Lianghao Deng and Mei Li and Mingfeng Xue and Mingze Li and Pei Zhang and Peng Wang and Qin Zhu and Rui Men and Ruize Gao and Shixuan Liu and Shuang Luo and Tianhao Li and Tianyi Tang and Wenbiao Yin and Xingzhang Ren and Xinyu Wang and Xinyu Zhang and Xuancheng Ren and Yang Fan and Yang Su and Yichang Zhang and Yinger Zhang and Yu Wan and Yuqiong Liu and Zekun Wang and Zeyu Cui and Zhenru Zhang and Zhipeng Zhou and Zihan Qiu},
  year={2025}, eprint={2505.09388}, archivePrefix={arXiv}, primaryClass={cs.CL},
  url={https://arxiv.org/abs/2505.09388}
}

@misc{shao2024deepseekmath,
  title={{DeepSeekMath}: Pushing the Limits of Mathematical Reasoning in Open Language Models},
  author={Zhihong Shao and Peiyi Wang and Qihao Zhu and Runxin Xu and Junxiao Song and Xiao Bi and Haowei Zhang and Mingchuan Zhang and Y. K. Li and Y. Wu and Daya Guo},
  year={2024}, eprint={2402.03300}, archivePrefix={arXiv}, primaryClass={cs.CL},
  url={https://arxiv.org/abs/2402.03300}
}

@misc{qwen38,
  title={{Qwen3.8-27B} Model Card},
  author={{Qwen Team}},
  year={2026},
  url={https://huggingface.co/Qwen/Qwen3.8-27B}
}

@misc{lu2026activationconditionedselfdistillation,
  title={Activation-Conditioned Self-Distillation},
  author={Zhexi Lu and Subhajit Chaudhury and Tejaswini Pedapati and Keerthiram Murugesan and Lei Yu},
  year={2026},
  eprint={2609.38342},
  archivePrefix={arXiv},
  primaryClass={cs.LG},
  url={https://arxiv.org/abs/2609.38342}
}

@misc{openai2026gpt56,
  title = {{GPT-5.6} System Card},
  author = {{OpenAI}},
  year = {2026},
  month = {July},
  url = {https://deploymentsafety.openai.com/gpt-5-6}
}

@misc{liu2026whenteachertokensreliable,
  title={When Are Teacher Tokens Reliable? Position-Weighted On-Policy Self-Distillation for Reasoning},
  author={Xiaogeng Liu and Xinyan Wang and Yingzi Ma and Yechao Zhang and Chaowei Xiao},
  year={2026},
  eprint={2605.21606},
  archivePrefix={arXiv},
  primaryClass={cs.LG},
  url={https://arxiv.org/abs/2605.21606}
}

@misc{verga2024replacingjudgesjuriesevaluating,
  title={Replacing Judges with Juries: Evaluating {LLM} Generations with a Panel of Diverse Models},
  author={Pat Verga and Sebastian Hofstatter and Sophia Althammer and Yixuan Su and Aleksandra Piktus and Arkady Arkhangorodsky and Minjie Xu and Naomi White and Patrick Lewis},
  year={2024},
  eprint={2404.18796},
  archivePrefix={arXiv},
  primaryClass={cs.CL},
  url={https://arxiv.org/abs/2404.18796}
}

@misc{xie2026smalljudgesrubricrl,
  title={Small Language Models as Judges for Rubric-Based Reinforcement Learning},
  author={Fengyu Xie and Yilun Zhao and Bingsen Chen and Arman Cohan and Chen Zhao},
  year={2026},
  eprint={2608.30005},
  archivePrefix={arXiv},
  primaryClass={cs.CL},
  url={https://arxiv.org/abs/2608.30005}
}

\appendix

\section{Related Work}
\label{sec:related-extended}

\paragraph{Distillation and self-training.}
Knowledge distillation transfers teacher predictions to a student \citep{hinton2015distillingknowledgeneuralnetwork,kim-rush-2016-sequence}; on-policy distillation applies this guidance along student-generated responses \citep{agarwal2024onpolicydistillationlanguagemodels}. Context distillation transfers behavior elicited by additional teacher context \citep{askell2021generallanguageassistantlaboratory,snell2022learningdistillingcontext,ye2026onpolicycontextdistillationlanguage}, connecting to learning with privileged information \citep{vapnik2009newlearningparadigmlearning,lopezpaz2016unifyingdistillationprivilegedinformation}. Self-training selects generated responses using consistency, correctness checks, or rewards \citep{huang-etal-2023-large,zelikman2022starbootstrappingreasoningreasoning,gulcehre2023reinforcedselftrainingrestlanguage,dong2023raftrewardrankedfinetuning,singh2024beyondhumandatascaling}. Self-Refine instead revises responses at inference time, whereas self-rewarding models construct preferences for training \citep{madaan2023selfrefineiterativerefinementselffeedback,yuan2024selfrewardinglanguagemodels}.

\paragraph{Sources of teacher guidance.}
SDFT uses demonstrations for continual learning, and OPSD uses reference solutions for reasoning \citep{shenfeld2026selfdistillationenablescontinuallearning,zhao2026selfdistilledreasoneronpolicyselfdistillation}. RCSD first trains a rubric generator through a teacher given reference answers, then supplies its rubrics to the reasoning teacher \citep{gu2026rethinkingrewardsupervisionrubricconditioned}. RGSD provides rubrics to a frozen teacher and trains a question-only student \citep{rezaei2026rubricguidedselfdistillationposttrainingrubric}. Experiential Learning obtains textual coaching feedback for non-verifiable tasks \citep{ye2026llmasacoachexperientiallearningnonverifiable}. Recent analyses find that OPSD's gains need not depend on the paired reference solution, because solutions to other problems can provide comparable teacher context \citep{ichihara2026privilegedsolutionscontextinducedteacher,shrestha2026rethinkingprivilegedinformationonpolicy}. These sources of teacher guidance differ from \method{}'s use of the completed attempt itself as additional teacher context.

\paragraph{Attempts, feedback, and reference-free supervision.}
SDPO constructs teacher context from textual feedback, or successful responses when only scalar rewards are available \citep{hubotter2026reinforcementlearningselfdistillation}. H$^2$SD combines verified successful attempts with reference hints for failures, while PAST uses complete attempts, reference solutions, and correctness-based routing \citep{cai2026h2sdhybridhindsightselfdistillation,feng2026pastprivilegedadaptationcomplete}. SD-Zero trains a reviser using binary correctness feedback in mathematics and coding, then distills its predictions into the generator \citep{he2026selfdistillationzeroselfrevisionturns}. Activation-Conditioned Self-Distillation uses verified outcomes to construct teacher-steering activation contrasts \citep{lu2026activationconditionedselfdistillation}. OASIS conditions the teacher on another unverified attempt and uses final-answer labels \citep{hossain2026overcomingscalinglimitsonpolicy}, and conditioning the teacher on a sampled correct response can reduce output diversity \citep{nicolicioiu2026onpolicyselfdistillationsampleddemonstrations}. Without supplied solutions, Canon and u-OPSD condition the teacher on a self-generated solution that reaches the majority answer over multiple samples; CoDA additionally penalizes minority trajectories \citep{gkountouras2026consensusprivilegedcontextlabelfree,li2026onpolicyselfdistillationanysupervision,guo2026learningconsensusdisagreementunsupervised}. NSD suppresses tokens favored by negative conditions generated from its own responses \citep{pei2026negativeselfdistillationlearningreason}. Unlike these methods, \method{} uses the completed attempt itself as additional teacher context, without correctness feedback, a reference solution, or answer consensus, and weights that attempt by judged reliability and student uncertainty.

\paragraph{Confidence and uncertainty.}
Confidence and uncertainty guide pseudo-label selection and weighting \citep{chen2023softmatchaddressingquantityqualitytradeoff,mukherjee2020uncertaintyawareselftrainingfewshot}. LLKD's weighted variant adds normalized teacher confidence and student uncertainty to scale selected classification losses \citep{li-etal-2025-learning}. For reasoning, EGRSD attenuates token updates using teacher entropy, while COSE weights PPO updates and prioritizes replay using evaluator confidence \citep{ke2026respectingselfuncertaintyonpolicyselfdistillation,wei2026confidenceorchestratedselfevolutionuncertainllm}. PW-OPSD weights reference-conditioned distillation by token position, motivated by diagnostics of teacher-token reliability \citep{liu2026whenteachertokensreliable}. Other approaches optimize predictive confidence through entropy minimization or confidence rewards \citep{agarwal2025unreasonableeffectivenessentropyminimization,prabhudesai2025maximizingconfidencealoneimproves,zhao2026learningreasonexternalrewards}. \method{} distinguishes confidence in judging an attempt's quality from uncertainty in predicting its tokens; evaluator bias and calibration remain relevant to interpreting both signals \citep{zheng2023judgingllmasajudgemtbench,panickssery2024llmevaluatorsrecognizefavor,guo2017calibrationmodernneuralnetworks}.

\paragraph{Scope of comparison.}
Our experiments compare representative supervision strategies on fixed NaturalReasoning questions. RGSD shares OPSD's teacher-conditioning principle, using rubrics in place of reference solutions \citep{rezaei2026rubricguidedselfdistillationposttrainingrubric}. We include OPSD with NaturalReasoning's Llama-generated responses as teacher context and rubric-conditioned RCSD as representative baselines, without separately evaluating RGSD. Verification-based methods require correctness feedback, while answer consensus requires rules for matching free-form answers. LLKD selects and weights pseudo-labels in a fixed classification label space, and COSE jointly generates tasks and updates its agents. Entropy-minimization and self-certainty methods also require only questions \citep{agarwal2025unreasonableeffectivenessentropyminimization,zhao2026learningreasonexternalrewards}; because student entropy decreases during \method{} training, they are a natural control that we leave to future comparison. Other demonstration-, rubric-, feedback-, and uncertainty-based methods remain applicable or adaptable alternatives outside the present comparison.

\section{Implementation Details}
\label{sec:implementation}

\begin{algorithm}[!htbp]
\caption{\method{} training procedure}
\label{alg:acsd}
\small
\begin{algorithmic}[1]
\Require Questions $\mathcal D$, initial model $\pi_{\theta_0}$, coefficient $\alpha$
\Ensure Trained reasoning model $\pi_\theta$
\State Initialize $\theta\gets\theta_0$; retain $\pi_{\theta_0}$ as judge
\For{each sampled question $\mathbf{x}_i\in\mathcal D$}
    \State Generate attempt $\mathbf{y}_i$ using current $\pi_\theta$ (Eq.~\ref{eq:rollout})
    \State Compute student predictions $\pi^{\mathrm S}_{i,t}$ (Eq.~\ref{eq:aligned})
    \State Compute teacher predictions $\pi^{\mathrm T}_{i,t}$ (Eq.~\ref{eq:aligned})
    \State Obtain $\bar s_i$ and $c_i^{\mathrm J}$ (Eqs.~\ref{eq:judge-probs}--\ref{eq:judge-confidence})
    \State Compute judged reliability $r_i^{\mathrm J}$ (Eq.~\ref{eq:quality})
    \State Compute student uncertainty $u_i^{\mathrm S}$ (Eq.~\ref{eq:uncertainty})
    \State Form weight $w_i$ (Eq.~\ref{eq:weight})
    \State Update $\theta$ (Section~\ref{sec:loss})
\EndFor
\State \Return $\pi_\theta$
\end{algorithmic}
\end{algorithm}

\paragraph{Prompts.}
Table~\ref{tab:prompts} lists the three prompts. Student and teacher prompts use the model's thinking mode, and the judge prompt does not. The teacher and judge receive the same decoded attempt.

\begin{table}[!h]
\centering
\small
\begin{tabular}{@{}p{0.97\columnwidth}@{}}
\toprule
\textbf{Student} (user turn)\\
\texttt{Question: \{question\}}\\
\texttt{Please reason step by step, and put your final answer within \textbackslash boxed\{\}.}\\
\midrule
\textbf{Teacher} (user turn)\\
\texttt{Question: \{question\}}\\
\texttt{Here is a candidate analysis that may contain useful or misleading information:}\\
\texttt{=== Candidate Analysis Begin === \{attempt\} === Candidate Analysis End ===}\\
\texttt{Solve the problem independently using any useful information from the candidate. Please reason step by step, and put your final answer within \textbackslash boxed\{\}.}\\
\midrule
\textbf{Judge} (system turn)\\
\texttt{You are an impartial evaluator of reasoning responses. Evaluate logical validity, internal consistency, factual plausibility, relevance, and completeness. Use this rubric strictly: 1 = fundamentally incorrect, irrelevant, or nonsensical; 2 = contains major errors that make the conclusion unreliable; 3 = partially correct but contains an important error, omission, or incomplete conclusion; 4 = mostly correct and complete, with only minor issues that do not invalidate the conclusion; 5 = fully correct, complete, relevant, and well justified. A response that omits a requested final answer, conclusion, or subpart is incomplete and must not receive a score above 3.}\\
\textbf{Judge} (user turn)\\
\texttt{Question: \{question\} Response: \{attempt\}}\\
\texttt{Apply the 1-to-5 rubric above. Return exactly one digit: 1, 2, 3, 4, or 5.}\\
\bottomrule
\end{tabular}
\caption{Student, attempt-conditioned teacher, and judge prompts used by \method{}. The judge applies a common five-level quality rubric.}
\label{tab:prompts}
\end{table}

\paragraph{Teacher and judge.}
Generation, teacher prediction, and student prediction use the same current LoRA adapters \citep{hu2021loralowrankadaptationlarge}. Gradients through student predictions update these shared adapters, so the teacher also uses the updated parameters on the next step. We refer to this as a dynamic teacher; a fixed teacher retains the initial parameters. The judge uses the initial backbone with adapters disabled and reads the five score-token logits after its prompt, without generating text. For each update, teacher predictions and response weights are computed without gradient tracking; generation and judging are also outside the differentiated loss.

\paragraph{Judged reliability example.}
A judge that splits its probability between ratings 4 and 5 as 0.44/0.56 assigns score 5 ($\bar s_i=1$) but has confidence of only about 0.57, so $r_i^{\mathrm J}\approx0.57$; a decisive rating of 5 gives reliability close to 1.

\paragraph{Prediction settings.}
The generation notation in Equation~\ref{eq:rollout} denotes one attempt from $\pi_\theta$ using the temperature and top-$p$/top-$k$ settings in Table~\ref{tab:hyperparameters}. Student and teacher predictions in Equation~\ref{eq:aligned} instead use a full-vocabulary softmax with logits divided by the distillation temperature. Student entropy uses this same full-vocabulary distribution, without the generation filters. The judge's scoring logits in Equation~\ref{eq:judge-probs} are its five rating-token logits divided by the judge temperature, which is 1.0 in the main configuration.

\paragraph{Loss implementation.}
We use the OPSD implementation \citep{zhao2026selfdistilledreasoneronpolicyselfdistillation}, which caps each teacher-weighted log-probability-ratio contribution at $\kappa=0.05$ before summing over vocabulary coordinates. This is a coordinate-clipped surrogate of the forward-KL objective in Equation~\ref{eq:loss}, rather than exact KL; the cap can change the update direction \citep{hossain2026overcomingscalinglimitsonpolicy,huang2026whenclippingreversescorrection}. Distillation and student entropy use the full output vocabulary without generation sampling filters.

\paragraph{Training configuration.}
Table~\ref{tab:hyperparameters} collects the main \method{} settings for both model sizes. We use linear learning-rate decay. The effective batch size is 32 questions, combining one question per GPU with eight gradient accumulation steps across four GPUs. The epoch budget is a configured maximum; the main results are reported after 150 training steps. The coefficient study varies the response-weight setting with the same dynamic teacher used in the main configuration. One \method{} training run took approximately 2 days on four NVIDIA A100 80\,GB GPUs.

\begin{table}[t]
\centering

\small
\begin{tabular}{@{}l r@{}}
\toprule
\textbf{Parameter} & \textbf{Setting}\\
\midrule
Optimizer & AdamW\\
Learning rate & $5\times10^{-6}$\\
Epoch budget (maximum) & 3\\
Effective batch size & 32\\
Training steps for reported results & 150\\
\midrule
Attempts per sampled question & 1\\
Generation temperature & 1.2\\
Sampling top-$p$ / top-$k$ & 0.95 / 20\\
Max. response length (tokens) & 8,192\\
Max. sequence length (tokens) & 32,768\\
\midrule
Distillation temperature $\tau_{\mathrm D}$ & 1.2\\
Vocabulary-coordinate cap $\kappa$ & 0.05\\
\bottomrule
\end{tabular}
\caption{Key training hyperparameters for the main Qwen3-4B and Qwen3-8B \method{} runs.}
\label{tab:hyperparameters}
\end{table}

\paragraph{Masks and reduction.}
The implementation excludes prompt and padding tokens from entropy and loss calculations. Equations~\ref{eq:uncertainty} and~\ref{eq:loss} sum over the unpadded response positions of retained responses. Responses that reach the 8,192-token generation limit are excluded from distillation and receive zero weight. A response with no valid tokens contributes no loss and has $u^{\mathrm S}_i=0$. The main configuration uses one question per device microbatch, so the valid-token-normalized loss is the per-response average in Equation~\ref{eq:loss}. The implementation sums weighted token losses and divides by the microbatch's valid-token count before gradient accumulation and distributed averaging. With larger microbatches, this is token-count weighting rather than an equal average of response means. Algorithm~\ref{alg:acsd} presents the per-question computation; optimizer updates follow the accumulation schedule described above. No gradient passes through generation, judging, teacher predictions, or weights.

\paragraph{Data.}
We use the public 1.1M-question release of NaturalReasoning \citep{yuan2025naturalreasoningreasoningwild28m}. Each row pairs a question with a reference answer derived from its source document, when one exists, and a response generated by Llama3.3-70B-Instruct. All methods we train share one selection: we keep rows whose reference-answer field is non-empty, shuffle with seed 42, and take the first 30,000 questions. In \method{}, the reference field is used only for this selection, and neither dataset answer field is used in training prompts. The student prompt contains the question and task instructions; teacher and judge prompts also include the model's own completed attempt.

\paragraph{GRPO reward.}
For GRPO, scalar response-quality rewards are supplied by a frozen external Qwen3.8-27B judge \citep{qwen38}.

\paragraph{Training steps for reported results.}
At both Qwen3-4B and Qwen3-8B, the main SFT, RCSD, OPSD, NSD, and \method{} results are reported after 150 training steps, and GRPO results after 500 training steps. Training steps denote optimizer updates. For \method{}, the reporting step was fixed at 150 independently of the checkpoint evaluations. The additional evaluations in Appendix~\ref{sec:training-duration} characterize performance over training.

\paragraph{Evaluation decoding.}
We use temperature 1.0, top-$p=0.95$, top-$k=-1$ (disabled), min-$p=0$, and presence penalty 0. We enable thinking mode and generate four responses per question. These evaluation settings differ from the training generation settings in Table~\ref{tab:hyperparameters}.

\paragraph{Benchmark scoring.}
Benchmark-specific automatic scoring compares parsed multiple-choice answers with annotated options and extracted mathematical answers with references using numerical or symbolic comparison. ZebraLogic requires every grid assignment to be correct. These scores use neither the training judge nor an external LLM judge.

\paragraph{Training-dynamics summaries and intervals.}
Figures~\ref{fig:dynamics}(b--c) and~\ref{fig:student-groups} use the same reported summaries of completed score-4/5 responses. Figure~\ref{fig:dynamics}(a) uses the corresponding saved summaries at steps 100, 200, and 300, including all completed responses with scores 1--5. For Figure~\ref{fig:dynamics}(b), the denominator is all responses with the corresponding judge score, and the numerator counts high-confidence judgments. For Figure~\ref{fig:student-groups}(b), $n$ is the judge-group response count and $k$ counts responses with student entropy below the pooled mean of the corresponding summary. Both proportion plots use 95\% Wilson confidence intervals with $z=1.96$. For example, the score-5/high-confidence group at step 100 has $k=231$ and $n=358$, giving 64.5\% and an interval of 59.4--69.3\%; at step 300, $463/638$ gives 72.6\% with an interval of 69.0--75.9\%. These intervals describe response proportions, rather than variation across independent training runs. The band in Figure~\ref{fig:dynamics}(c) instead spans the 25th--75th percentiles of response entropy; it is not a confidence interval.

\paragraph{External quality audit.}
We randomly sample 200 saved question--response pairs and ask \texttt{gpt-5.6sol} \citep{openai2026gpt56} to act as an external judge using the same five-level quality rubric in Table~\ref{tab:prompts}. Each of the 200 questions contributes one response and one observation. The three regressions include an intercept and predict the external GPT rating from the training judge's score, confidence, or both. We report descriptive in-sample fit and coefficient estimates. The multiple correlation coefficient $R=\sqrt{R^2}$ is the correlation between fitted and observed GPT ratings. Its square, $R^2$, is the fraction of external-rating variation explained within the fitted sample. $R$ is nonnegative; the signs of the regression coefficients give the direction of the fitted conditional associations. The joint coefficients are $b_{\mathrm{score}}=0.84$ per rating level and $b_{\mathrm{confidence}}=3.78$. Thus, a 0.1 increase in confidence is associated with a 0.378 increase in the external rating at the same judge score; an increase of 1 in the judge score is associated with a 0.84 increase in the rating at the same confidence. These estimates describe response-quality ratings, rather than verified correctness or calibrated probabilities.

\section{Additional Experimental Results}
\label{sec:extended-results}

\subsection{Extended weighting results}
\label{sec:extended-weighting}
\begin{table}[!htbp]
\centering
\small
\begin{tabular}{@{}l r r r@{}}
\toprule
\textbf{Benchmark} & \textbf{Score only} & \textbf{\method{}} & $\Delta$\\
\midrule
GPQA Diamond & 52.40 & \textbf{52.78} & +0.38\\
SciBench & 66.11 & \textbf{70.66} & +4.55\\
MMLU & 78.12 & \textbf{79.38} & +1.26\\
LiveBench Reasoning & 58.50 & \textbf{61.25} & +2.75\\
OpenBookQA & 91.75 & \textbf{92.60} & +0.85\\
GSM8K & 94.56 & \textbf{94.81} & +0.25\\
MATH & 83.47 & \textbf{85.87} & +2.40\\
AMC & 71.39 & \textbf{74.10} & +2.71\\
Minerva Math & 32.08 & \textbf{35.02} & +2.94\\
OlympiadBench & 51.41 & \textbf{54.27} & +2.86\\
ZebraLogic & 73.02 & \textbf{74.38} & +1.36\\
\midrule
Macroaverage & 68.44 & \textbf{70.47} & +2.03\\
\bottomrule
\end{tabular}
\caption{Qwen3-4B accuracy (Avg@4, \%) under score-only and \method{} weighting. $\Delta$ is \method{}'s absolute gain over score-only weighting on the same 0--100 scale. \method{} improves all eleven benchmarks.}
\label{tab:weight-full}
\end{table}

Table~\ref{tab:weight-full} reports the complete benchmark comparison between score-only weighting and \method{}. The full rule is higher on each benchmark, including scientific reasoning, mathematics, and logic-grid reasoning, with single-run differences below 1 on several benchmarks. This comparison changes confidence and student uncertainty together; the endpoint comparisons in Section~\ref{sec:ablations} examine reliability-only and uncertainty-only weighting.

\paragraph{Sensitivity to SciBench.}
The 4B gain is largest on SciBench, where the base model starts at 35.77\% and \method{} reaches 70.66\%. Removing this benchmark from the macroaverage still leaves \method{} ahead of the base model by 6.30 and the strongest remaining baseline, OPSD, by 3.61. Thus, the overall advantage is not confined to SciBench.

\subsection{Performance across training steps}
\label{sec:training-duration}

\begin{figure}[!tbp]
\centering
\includegraphics[width=\columnwidth]{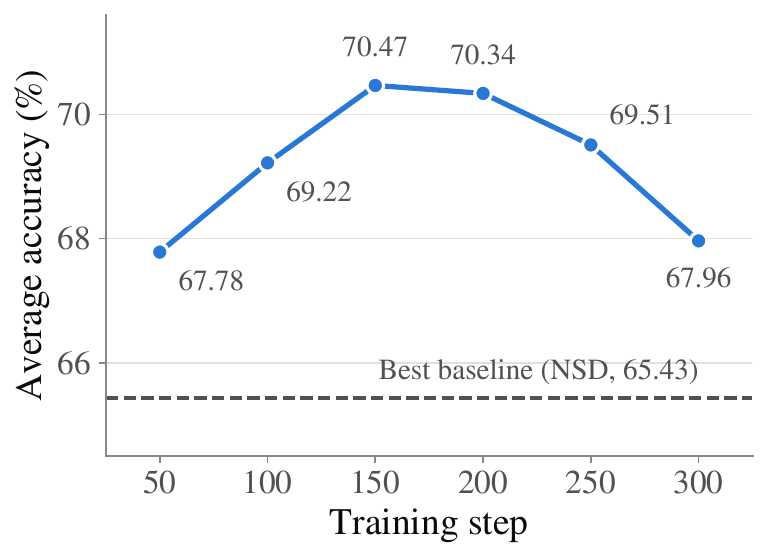}
\caption{Average accuracy (Avg@4, \%) across eleven benchmarks for Qwen3-4B \method{} with $\alpha=0.5$ at six evaluated training steps from 50 to 300. The highest observed value is 70.47\% at step 150. The dashed line marks the strongest Qwen3-4B baseline average (NSD, 65.43\%; Table~\ref{tab:main}).}
\label{fig:training-steps}
\end{figure}

Accuracy rises from 67.78\% at step 50 to 70.47\% at step 150, remains close at step 200 (70.34\%), and falls to 67.96\% at step 300 (Figure~\ref{fig:training-steps}). Figure~\ref{fig:dynamics}(c) shows that student entropy continues to fall as this accuracy curve declines. Entropy measures predictive uncertainty, whereas benchmark accuracy measures task performance. Their divergence shows why the learning-priority signals should be interpreted alongside the benchmark curve. Training step 150 gives the highest observed reasoning performance, rather than the lowest student entropy, and every evaluated training step exceeds the strongest baseline, NSD (65.43\%).

\section{Retaining a Correct Solution with Increasing Certainty}
\label{sec:case-study}

Table~\ref{tab:case-study} reproduces the complete saved final responses to the same heat-calculation question with $\alpha=0.5$. They are observations at training steps 30 and 105, rather than responses regenerated for separate evaluations at those training steps. Both responses contain the correct two-stage calculation. Their shared correctness makes this case useful for examining changes that the discrete judge score alone does not capture. We compare the logged training signals and verify the saved final calculations below.

\begin{table*}[p]
\centering
\fontsize{9}{10.3}\selectfont
\setlength{\smallskipamount}{0pt}
\setlength{\abovedisplayskip}{1pt}
\setlength{\belowdisplayskip}{2pt}
\setlength{\abovedisplayshortskip}{2pt}
\setlength{\belowdisplayshortskip}{2pt}
\begin{tabular}{@{}>{\raggedright\arraybackslash}p{0.485\textwidth}@{\hspace{0.03\textwidth}}>{\raggedright\arraybackslash}p{0.485\textwidth}@{}}
\toprule
\multicolumn{2}{@{}p{\textwidth}@{}}{\textbf{Question.} A solid has a melting point of $90.0^\circ$C. Calculate the total heat required to change 2.5 kg of this solid from $40.0^\circ$C to a liquid, given that the specific heat of the solid is 390 J/kg-K and its heat of fusion is 4000 J/kg. Show your work and explain your reasoning.}\tabularnewline[4pt]
\midrule
\textbf{Earlier response (step 30)} & \textbf{Later response (step 105)}\tabularnewline[3pt]
To determine the total heat required to change 2.5 kg of a solid from 40.0$^\circ$C to a liquid, we need to account for two distinct processes:
\par\smallskip
\noindent 1. \textbf{Heating the solid from its initial temperature to its melting point}\par
\noindent 2. \textbf{Melting the solid at its melting point into a liquid}\par
&
To calculate the total heat required to change 2.5 kg of a solid from 40.0$^\circ$C to a liquid, we need to consider two distinct processes:
\par\smallskip
\noindent 1. \textbf{Heating the solid from its initial temperature (40.0$^\circ$C) to its melting point (90.0$^\circ$C).}\par
\noindent 2. \textbf{Melting the solid at its melting point (90.0$^\circ$C) into a liquid.}\par\tabularnewline

\addlinespace[4pt]
\textbf{Step 1: Heating the solid to its melting point}
\par\smallskip
The temperature difference between the initial temperature and the melting point is:
\par\smallskip

\[
\Delta T = 90.0^\circ C - 40.0^\circ C = 50.0^\circ C
\]

\par\smallskip
The heat required to raise the temperature of the solid is calculated using the specific heat formula:
\par\smallskip

\[
Q_1 = m \cdot c \cdot \Delta T
\]

\par\smallskip
Where:\par
\noindent\(\bullet\) $ m = 2.5 \, \text{kg} $\par
\noindent\(\bullet\) $ c = 390 \, \text{J/kg}\mathord{\cdot}\text{K} $\par
\noindent\(\bullet\) $ \Delta T = 50.0 \, \text{K} $ (since a change of 1$^\circ$C = 1 K)\par
\par\smallskip

\[
\begin{aligned}Q_1 &= 2.5 \, \text{kg} \cdot 390 \, \text{J/kg}\mathord{\cdot}\text{K} \cdot 50.0 \, \text{K}\\ &= 48750 \, \text{J}\end{aligned}
\]

&
\textbf{Step 1: Heat Required to Raise the Temperature of the Solid}
\par\smallskip
The formula for the heat required to change the temperature of a substance is:
\par\smallskip

\[
Q_1 = m \cdot c \cdot \Delta T
\]

\par\smallskip
Where:\par
\noindent\(\bullet\) $ m = 2.5 \, \text{kg} $ (mass of the solid),\par
\noindent\(\bullet\) $ c = 390 \, \text{J/kg-K} $ (specific heat of the solid),\par
\noindent\(\bullet\) $\Delta T = T_{\text{melt}} - T_{\text{initial}}$\par\hspace{1em}$=90.0^\circ\text{C}-40.0^\circ\text{C}=50.0^\circ\text{C}$.\par
\par\smallskip

\[
\begin{aligned}Q_1 &= 2.5 \, \text{kg} \times 390 \, \text{J/kg-K} \times 50.0 \, \text{K}\\ &= 48,750 \, \text{J}\end{aligned}
\]
\tabularnewline

\addlinespace[4pt]
\textbf{Step 2: Melting the solid at its melting point}
\par\smallskip
The heat required to melt the solid is calculated using the heat of fusion formula:
\par\smallskip

\[
Q_2 = m \cdot L_f
\]

\par\smallskip
Where:\par
\noindent\(\bullet\) $ m = 2.5 \, \text{kg} $\par
\noindent\(\bullet\) $ L_f = 4000 \, \text{J/kg} $\par
\par\smallskip

\[
\begin{aligned}Q_2 &= 2.5 \, \text{kg} \cdot 4000 \, \text{J/kg}\\ &= 10000 \, \text{J}\end{aligned}
\]

&
\textbf{Step 2: Heat Required to Melt the Solid}
\par\smallskip
The formula for the heat required to melt a substance is:
\par\smallskip

\[
Q_2 = m \cdot L_f
\]

\par\smallskip
Where:\par
\noindent\(\bullet\) $ m = 2.5 \, \text{kg} $ (mass of the solid),\par
\noindent\(\bullet\) $ L_f = 4000 \, \text{J/kg} $ (heat of fusion).\par
\par\smallskip

\[
\begin{aligned}Q_2 &= 2.5 \, \text{kg} \times 4000 \, \text{J/kg}\\ &= 10,000 \, \text{J}\end{aligned}
\]
\tabularnewline

\addlinespace[4pt]
\textbf{Step 3: Total Heat Required}
\par\smallskip

\[
\begin{aligned}Q_{\text{total}} &= Q_1 + Q_2\\ &= 48750 \, \text{J} + 10000 \, \text{J}\\ &= 58750 \, \text{J}\end{aligned}
\]

&
\textbf{Step 3: Total Heat Required}
\par\smallskip

\[
\begin{aligned}Q_{\text{total}} &= Q_1 + Q_2\\ &= 48,750 \, \text{J} + 10,000 \, \text{J}\\ &= \boxed{58750} \, \text{J}\end{aligned}
\]
\tabularnewline

\addlinespace[4pt]
\textbf{Final Answer}
\par\smallskip

\[
\boxed{58750}
\]

&
\textbf{Final Answer}
\par\smallskip

\[
\boxed{58750} \, \text{J}
\]
\tabularnewline
\bottomrule
\end{tabular}
\par\vspace{5pt}
\begin{tabular}{@{}l c c@{}}
\toprule
\textbf{Recorded training signal} & \textbf{Step 30} & \textbf{Step 105}\\
\midrule
Judge score $s_i$ & 5 & 5\\
Judge probabilities $\pi_i^{\mathrm J}(4)$ / $\pi_i^{\mathrm J}(5)$ & 0.438 / 0.562 & 0.119 / 0.879\\
Judge confidence $c_i^{\mathrm J}$ & 0.572 & 0.771\\
Student uncertainty $u_i^{\mathrm S}$ & 0.1728 & 0.0388\\
\bottomrule
\end{tabular}
\caption{Complete saved final responses (after thinking) and training signals for a selected heat-calculation case from an auxiliary fixed-teacher run. Both answers are correct; judge confidence increases and student uncertainty decreases between steps 30 and 105. Appendix~\ref{sec:case-study} gives the calculations and case-selection details.}
\label{tab:case-study}
\end{table*}

\paragraph{Checking both solutions.}
The solid must first reach its melting point and then undergo the phase change. Heating requires $mc\Delta T=2.5\times390\times(90-40)=48{,}750$\,J. Melting requires $mL_f=2.5\times4{,}000=10{,}000$\,J. The total is therefore $58{,}750$\,J, matching the supplied reference answer. Both saved final responses explicitly apply these formulas and obtain these values. A temperature difference of $50\,^{\circ}\mathrm C$ equals $50$\,K, so their temperature conversion is also correct. This check concerns the saved final explanations shown in the case record.

\paragraph{What changes when correctness is retained?}
The judge score remains 5, while the rating distribution becomes more concentrated: the probabilities assigned to scores 4 and 5 change from approximately 0.438/0.562 to 0.119/0.879. Judge confidence consequently rises from 0.572 to 0.771. Meanwhile, mean normalized student entropy falls from 0.1728 to 0.0388, approximately a 4.5-fold reduction. The later response preserves the same correct reasoning with less predictive uncertainty; it also states the energy unit next to its final boxed value. The fixed score thus conceals two changes with different effects on the additive weight: greater judged reliability raises its reliability contribution, while lower student uncertainty reduces its uncertainty contribution. This is the response-level pattern motivating learning from a correct attempt whose generation and assessment can still become more certain.

\paragraph{Selection and interpretation.}
The case was selected post hoc through a non-blinded review of repeated training observations. Reference answers were used for this selection and the subsequent correctness check, not for training supervision. The displayed case was chosen after checking its full saved final explanations for correctness at both observations and requiring an unchanged judge score.

The case illustrates retained correctness with increasing certainty; it neither identifies an initial random guess nor isolates the cause of the change.

\section{Additional Training-Dynamics Analysis}
\label{sec:dynamics-counts}

\begin{table}[!htbp]
\centering
\small
\setlength{\tabcolsep}{3pt}
\begin{tabular}{rcccc}
\toprule
& \multicolumn{2}{c}{Judge score 5} & \multicolumn{2}{c}{Judge score 4} \\
\cmidrule(lr){2-3}\cmidrule(lr){4-5}
Step & High conf. & Low conf. & High conf. & Low conf. \\
\midrule
0 & 135/314 & 103/334 & 116/137 & 215/351 \\
50 & 156/319 & 165/332 & 109/196 & 204/359 \\
100 & 231/358 & 158/322 & 140/259 & 146/310 \\
150 & 335/463 & 107/311 & 151/249 & 135/314 \\
200 & 361/482 & 115/290 & 235/373 & 95/218 \\
250 & 438/586 & 124/295 & 216/338 & 98/200 \\
300 & 463/638 & 117/294 & 212/344 & 106/214 \\
\bottomrule
\end{tabular}
\caption{Response counts underlying Figures~\ref{fig:dynamics}(b), \ref{fig:student-groups}, and~\ref{fig:low-entropy-composition}. Each entry is $k/n$: $n$ is the group size and $k$ counts responses with student entropy below the pooled mean of completed score-4/5 responses in the same reported summary. High judge confidence means $c^{\mathrm J}\geq0.7$; low confidence means $c^{\mathrm J}<0.7$. The figures show reported steps 50--300; step 0 is included here for completeness.}
\label{tab:dynamics-counts}
\end{table}

Table~\ref{tab:dynamics-counts} gives the shared counts used in the training-summary analysis. Figure~\ref{fig:dynamics}(b) divides the high-confidence group size by all responses with the same judge score. Figure~\ref{fig:student-groups}(a) divides each group size by all completed score-4/5 responses, while Figure~\ref{fig:student-groups}(b) reports $k/n$ for the two high-confidence groups. For the score-5/high-confidence group at step 300, these proportions are $638/932=68.5\%$, $638/1490=42.8\%$, and $463/638=72.6\%$, respectively. Thus, the panels describe different proportions of the same response population. Figure~\ref{fig:dynamics}(a) uses the same saved summaries at steps 100, 200, and 300 but includes all completed responses: 1,254, 1,369, and 1,502, respectively. Their score-4/5 subsets contain 1,249, 1,363, and 1,490 responses, matching Table~\ref{tab:dynamics-counts}.

\paragraph{Score distributions and group composition.}
The score-5 share of completed responses rises from 54.2\% at reported step 100 to 62.1\% at step 300 (Figure~\ref{fig:dynamics}a). Among completed score-4/5 responses, the score-5/high-confidence share grows from 26.5\% at reported step 50 to 42.8\% at step 300. The score-4/high-confidence share grows from 16.3\% to 23.1\%, reaching its maximum of 27.4\% at step 200 (Figure~\ref{fig:student-groups}a). Within score 5, the high-confidence proportion continues to rise after the accuracy peak, from 59.8\% at step 150 to 68.5\% at step 300.

\begin{figure}[!htbp]
\centering
\includegraphics[width=\columnwidth]{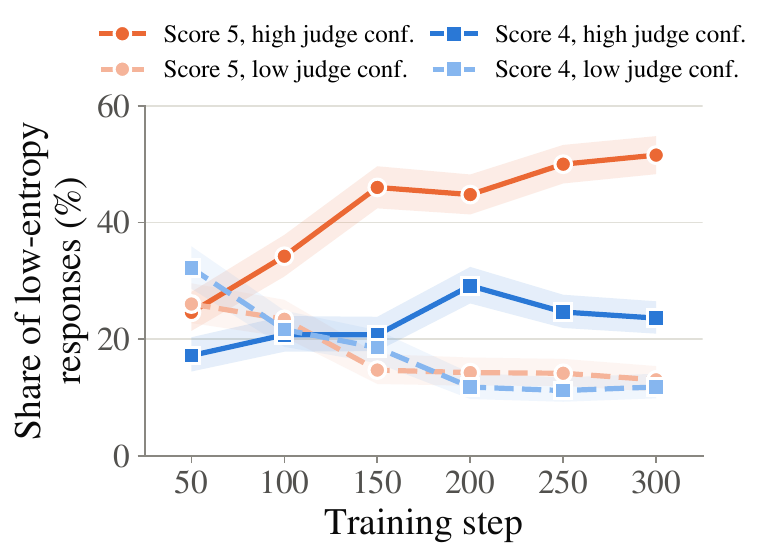}
\caption{Judge-group composition among low-entropy responses during Qwen3-4B \method{} training with $\alpha=0.5$. Only completed score-4/5 responses are included; low entropy means student entropy below their pooled mean at that reporting step. Each curve shows a group's share of this subset; the four shares sum to 100\%. High judge confidence means $c_i^{\mathrm J}\geq0.7$. Shading shows 95\% Wilson confidence intervals.}
\label{fig:low-entropy-composition}
\end{figure}

\paragraph{Composition of low-entropy responses.}
Among responses below the pooled mean student entropy, the score-5/high-confidence group's share rises from 24.6\% at step 50 to 51.6\% at step 300 (Figure~\ref{fig:low-entropy-composition}). At step 300, this share is $463/898$, whereas the within-group fraction in Figure~\ref{fig:student-groups}(b) is $463/638=72.6\%$. The composition depends on both group size and the fraction of each group below the mean.

\paragraph{Spread of student uncertainty.}
The 25th--75th percentile range of response uncertainty shifts from 0.21032--0.31335 at reported step 50 to 0.04992--0.08053 at step 300 (Figure~\ref{fig:dynamics}c). These quantiles describe variation across completed score-4/5 responses in each summary.

\paragraph{Detailed relative-certainty comparisons.}
At step 50, the fractions below the pooled mean student uncertainty are 48.9\% for score-5/high-confidence responses and 52.6\% overall; at step 100, they are 64.5\% and 54.0\%. The group's fraction peaks at 74.9\% at step 200 before decreasing to 72.6\% at step 300, when the pooled fraction is 60.3\%. It exceeds the pooled percentage from step 100 onward. The score-4/high-confidence group's fraction remains at or above the pooled percentage throughout, with almost no difference at step 100. High- and low-confidence score-5 responses have similar fractions at step 50 (48.9\% versus 49.7\%), but differ at step 300 (72.6\% versus 39.8\%).

\end{document}